\documentclass[letterpaper]{article} 
\usepackage{iclr2020_conference,times}

\usepackage{hyperref}
\usepackage{url}

\title{Expected Information Maximization\\ Using the I-Projection for Mixture Density Estimation}

\author{Philipp Becker \\
Autonomous Learning Robots, KIT\\
Bosch Center for Artificial Intelligence \\
\textit{Correspondence to: philipp.becker@kit.edu}
\And
Oleg Arenz \\
Intelligent Autonomous Systems, TU Darmstadt
\And
Gerhard Neumann \\
Autonomous Learning Robots, KIT \\
Bosch Center for Artificial Intelligence \\ University of T\"ubingen
}

\usepackage{amssymb}
\usepackage{amsmath}
\usepackage{graphicx}
\usepackage{subcaption}
\usepackage{algorithm2e}

\usepackage{pgfplots}
\pgfplotsset{compat=1.14}
\usetikzlibrary{external}
\usetikzlibrary{pgfplots.groupplots}

\newcommand{\KL}[2]{\textrm{KL}\left( {#1}||{#2}\right)}
\newcommand{\cvec}[1]{\boldsymbol{\mathrm{#1}}}
\newcommand{\cmat}[1]{\boldsymbol{\mathrm{#1}}}
\newcommand{\old}[1]{{#1}_{t}}

\iclrfinalcopy 

\begin{document}

\maketitle

\begin{abstract}
Modelling highly multi-modal data is a challenging problem in machine learning. 
Most algorithms are based on maximizing the likelihood, which corresponds 
to the M(oment)-projection of the data distribution to the model distribution.
The M-projection forces the model to average over modes it cannot represent.
In contrast, the I(information)-projection ignores such modes in the data and concentrates on the modes the model can represent.
Such behavior is appealing whenever we deal with highly multi-modal data where modelling single modes correctly is more important than covering all the modes. 
Despite this advantage, the I-projection is rarely used in practice due to the lack of algorithms that can efficiently optimize it based on data.
In this work, we present a new algorithm called Expected Information Maximization (EIM) for computing the I-projection solely based on samples for general latent variable models, where we focus on Gaussian mixtures models and Gaussian mixtures of experts.
Our approach applies a variational upper bound to the I-projection objective which decomposes the original objective into single objectives for each mixture component as well as for the coefficients, allowing an efficient optimization. Similar to GANs, our approach employs discriminators but uses a more stable optimization procedure, using a tight upper bound.
We show that our algorithm is much more effective in computing the I-projection than recent GAN approaches and we illustrate the effectiveness of our approach for modelling multi-modal behavior on two pedestrian and traffic prediction datasets.  

\end{abstract}

\section{Introduction}
Learning the density of highly multi-modal distributions is a challenging machine learning problem relevant to many fields such as modelling human behavior \citep{Pentland1999}. 
Most common methods rely on maximizing the likelihood of the data.
It is well known that the maximum likelihood solution corresponds to computing the M(oment)-projection of the data distribution to the parametric model distribution \citep{bishop2006prml}.
Yet, the M-projection averages over multiple modes in case the model distribution is not rich enough to fully represent the data \citep{bishop2006prml}. 
This averaging effect can result in poor models, that put most of the probability mass in areas that are not covered by the data.  
The counterpart of the M-projection is the I(nformation)-projection.
The I-projection concentrates on the modes the model is able to represent and ignores the remaining ones. Hence, it does not suffer from the averaging effect \citep{bishop2006prml}.  

In this paper, we explore the I-projection for mixture models which are typically trained by maximizing the likelihood via expectation maximization (EM) \citep{dempster1977em}. 
Despite the richness of mixture models, the averaging problem remains as we typically do not know the correct number of modes and it is hard to identify all modes of the data correctly.  
By the use of the I-projection, our mixture models do not suffer from this averaging effect and can generate more realistic samples that are less distinguishable from the data. 
In this paper we concentrate on learning Gaussian mixture models and conditional Gaussian mixtures of experts \citep{jacobs1991emm} where the mean and covariance matrix are generated by deep neural networks. 

We propose Expected Information Maximization (EIM) \footnote{Code available at \url{https://github.com/pbecker93/ExpectedInformationMaximization}}, a novel approach capable of computing the I-projection between the model and the data.
By exploiting the structure of the I-projection, we can derive a variational upper bound objective, which was previously used in the context of variational inference  \citep{maaloe2016auxiliary, ranganath2016hierarchical, arenz2018vips}.  
In order to work with this upper bound objective based on samples, we use a discriminator to approximate the required density ratio, relating our approach to  GANs \citep{goodfellow2014gan, nowozin2016fgan, uehara2016bgan}. 
The discriminator also allows us to use additional discriminative features to improve model quality.  
In our experiments, we demonstrate that EIM is much more effective in computing the I-projection than recent GAN approaches. 
We apply EIM to a synthetic obstacle avoidance task, an inverse kinematic task of a redundant robot arm as well as a pedestrian and car prediction task using the Stanford Drone Dataset \citep{Robicquet2016} and a traffic dataset from the Next Generation Simulation program.

\begin{figure}[t]
\centering
\begin{minipage}[t]{0.19\textwidth}
\subcaption*{{\tiny{(a) Expert Data}}}
\includegraphics[width=\textwidth]{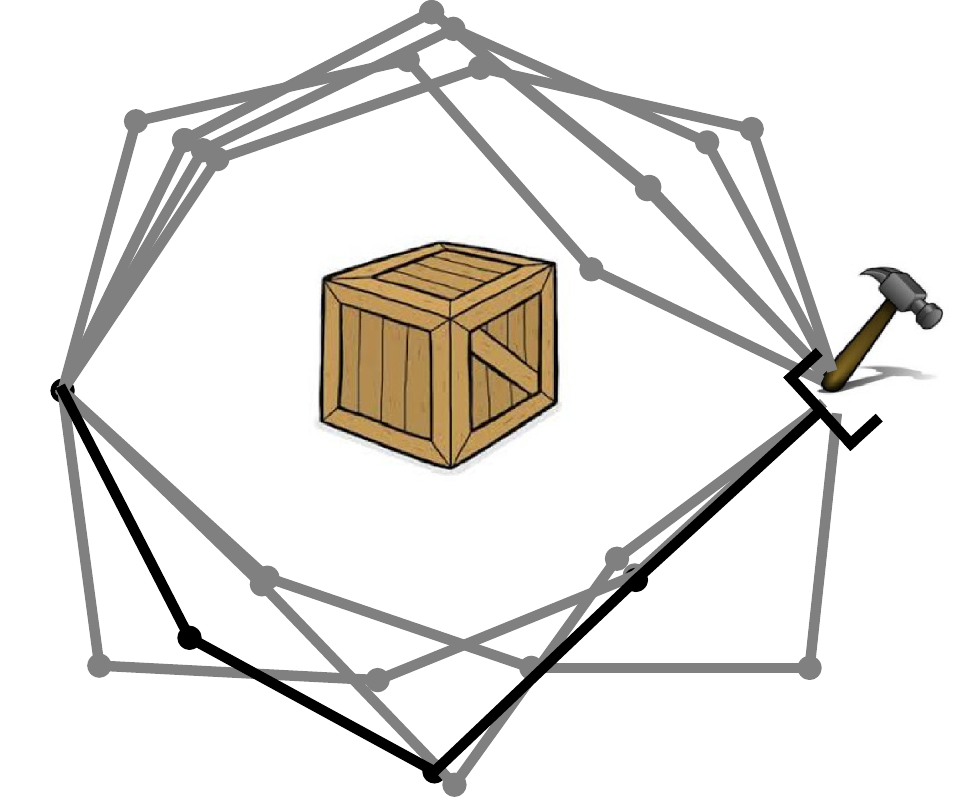}
\end{minipage}%
\hspace{0.07\textwidth}%
\begin{minipage}[t]{0.27\textwidth}
\subcaption*{\centering{\tiny{(b) M-projection}}}
\includegraphics[width=\textwidth]{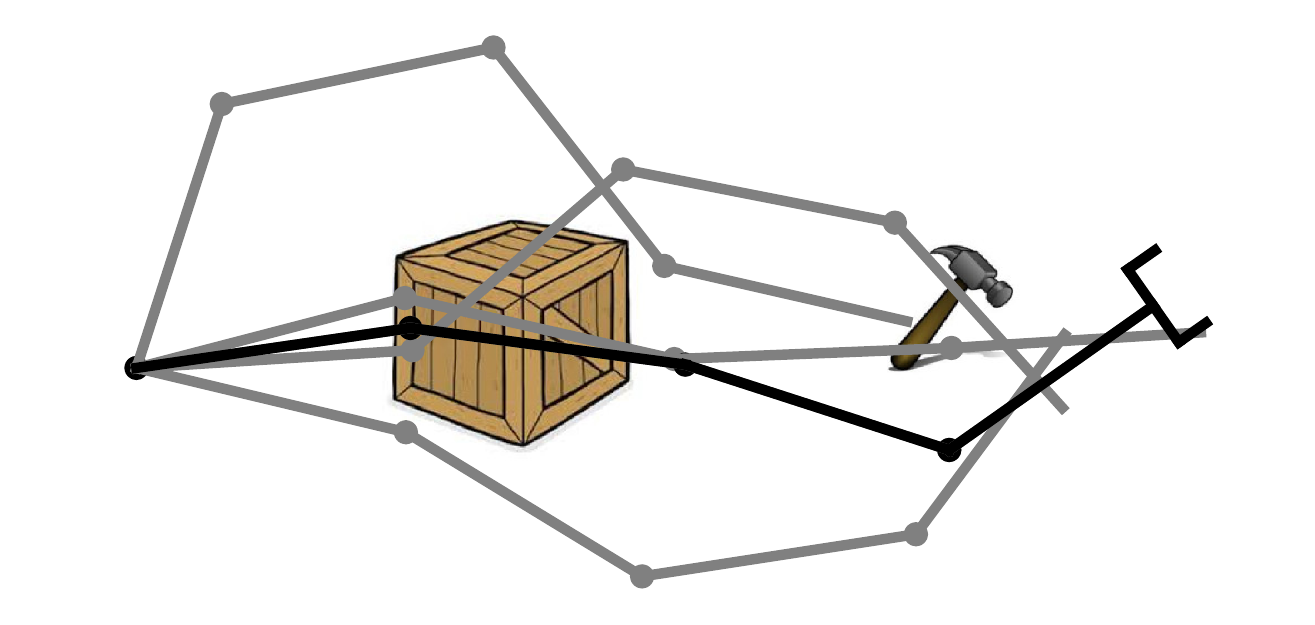}
\end{minipage}%
\hspace{0.07\textwidth}%
\begin{minipage}[t]{0.19\textwidth}
\subcaption*{\tiny{(c) I-projection}}
\includegraphics[width=\textwidth]{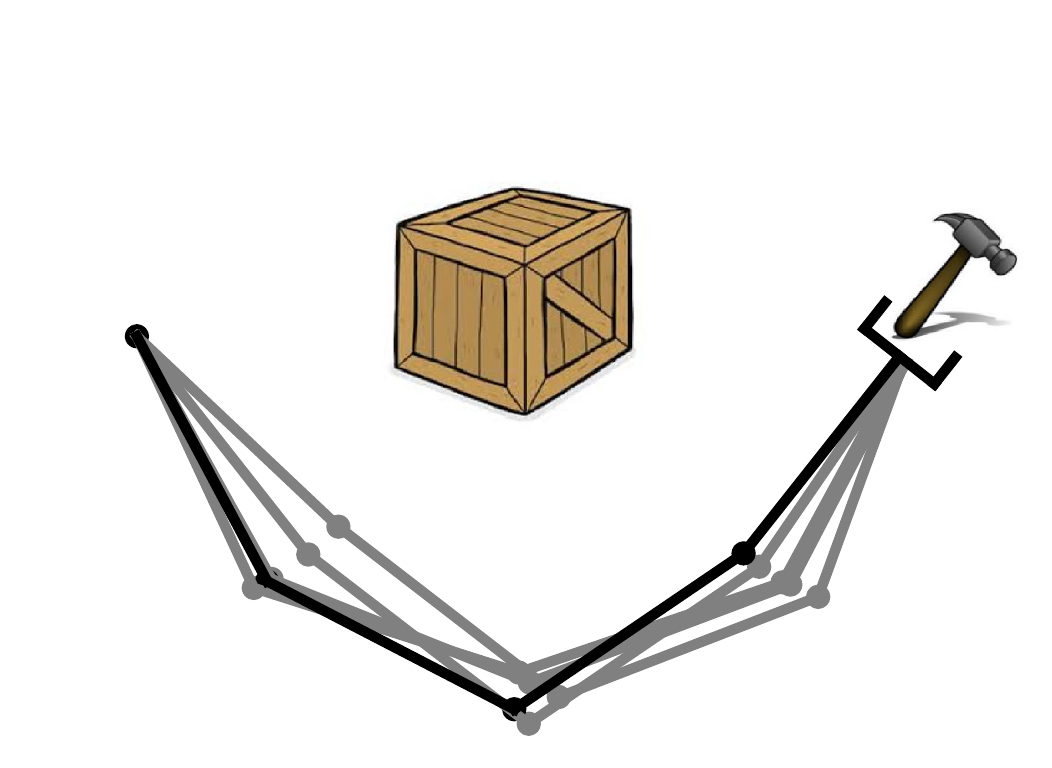}
\end{minipage}
\caption{Illustration of the I-projection vs. the M-projection for modelling behavior. (a): A robot reaches a target point while avoiding an obstacle. There are two different types of solutions, above and below the obstacle. A single Gaussian is fitted to the expert data in joint space. (b): The M-projection fails to reach the target and collides with the obstacle. (c): The I-projection ignores the second mode and reaches the target while avoiding the obstacle.}
\label{fig::robot_ex}
\end{figure}

\section{Preliminaries}

Our approach heavily builds on minimizing Kullback-Leibler divergences as well as the estimation of density ratios.
We will therefore briefly review both concepts.  

\paragraph{Density Ratio Estimation.}
\label{sec:dre}
Our approach relies on estimating density ratios $r(x) = q(x)  / p(x)$ based on samples of $q(x)$ and $p(x)$. \cite{sugiyama2012bregmandre} introduced a framework to estimate such density ratios based on the minimization of Bregman divergences \citep{bregman1967divegence}.
For our work we employ one approach from this framework, namely density ratio estimation by binary logistic regression. Assume a logistic regressor $C(x) = \sigma(\phi(x))$ with logits $\phi(x)$ and sigmoid activation function $\sigma$.
Further, we train $C(x)$ on predicting the probability that a given sample $x$ was sampled from $q(x)$. 
It can be shown that such a logistic regressor using a cross-entropy loss is optimal for $ C(x) = q(x) / \big(q(x) + p(x)\big).$
Using this relation, we can compute the log density ratio estimator by
\begin{align*}
 \log \frac{q(x)}{p(x)} = \log \frac{q(x) / \big(q(x) + p(x)\big)}{p(x) /\big(q(x) + p(x)\big)}
= \log \frac{C(x)}{1 - C(x)} = \sigma^{-1}\big(C(x)\big) = \phi(x).
\end{align*}
The logistic regressor is trained by minimizing the binary cross-entropy
\begin{align*}
    \textrm{argmin}_{\phi(x)} \textrm{BCE}(\phi(x), p(x), q(x)) = - \mathbb{E}_{q(x)} \left[\log \left(\sigma(\phi(x)) \right)\right] - \mathbb{E}_{p(x)} \left[\log \left(1- \sigma(\phi(x)) \right)  \right],
\end{align*}
where different regularization techniques such as $\ell_2$ regularization or dropout \citep{srivastava2014dropout} can be used to avoid overfitting. 

\paragraph{Moment and Information Projection.}
The Kullback-Leibler divergence \citep{kullback1951kl} is a standard similarity measure for distributions.
It is defined as 
$\KL{p(x)}{q(x)} =  \int p(x) \log p(x) / q(x) dx.$
Due to its asymmetry, the Kullback-Leibler Divergence provides two different optimization problems \citep{bishop2006prml} to fit a model distribution $q(x)$ to a target distribution  $p(x)$, namely 
$$\underbrace{\textrm{argmin}_{q(x)} \KL{p(x)}{q(x)}}_{\textrm{Moment-projection}} \quad \text{and} \quad \underbrace{\textrm{argmin}_{q(x)}  \KL{q(x)}{p(x)}}_{\textrm{Information-projection}}.$$
Here, we will assume that $p(x)$ is the data distribution, i.e., $p(x)$ is unknown but we have access to samples from $p(x)$. 
It can easily be seen that computing the M-projection to the data distribution is equivalent to maximizing the likelihood (ML) of the model \citep{bishop2006prml}. 
ML solutions match the moments of the model with the moments of the target distribution, which results in averaging over modes that can not be represented by the model.    
In contrast, the I-projection forces the learned generator $q(x)$ to have low probability whenever $p(x)$ has low probability, which is also called zero forcing. 

\section{Related Work}

We will now discuss competing methods for computing the I-projection that are based on GANs. 
Those are, to the best of our knowledge, the only other approaches capable of computing the I-projection solely based on samples of the target distribution. 
Furthermore, we will distinguish our approach from approaches based on variational inference that also use the I-projection.

\paragraph{Variational Inference.}
The I-projection is a common objective in Variational Inference \citep{opper2001meanField, bishop2006prml, kingma2013vae}. 
Those methods aim to fit tractable approximations to intractable distributions of which the unnormalized density is available.
EIM, on the other hand, does not assume access to the unnormalized density of the target distributions but only to samples.
Hence, it is not a variational inference approach, but a density estimation approach. 
However, our approach uses an upper bound that has been previously applied to variational inference \citep{maaloe2016auxiliary, ranganath2016hierarchical, arenz2018vips}. 
EIM is especially related to the VIPS algorithm \citep{arenz2018vips}, which we extend from the variational inference case to the density estimation case. 
Additionally, we introduce conditional latent variable models into the approach.

\paragraph{Generative Adversarial Networks.}
\label{sec:GAN}
While the original GAN approach minimizes the Jensen-Shannon Divergence \citep{goodfellow2014gan}, GANs have since been adopted to a variety of other distance measures between distributions, such as the Wasserstein distance \citep{arjovsky2017wasserstein}, symmetric KL \citep{chen2018symmetric} and arbitrary $f$-divergences \citep{ali1966fdiv, nowozin2016fgan, uehara2016bgan, poole2016iGenObj}.
Since the I-projection is a special case of an $f$-divergence, those approaches are of particular relevance to our work. 
\cite{nowozin2016fgan} use a variational bound for $f$-divergences \citep{nguyen2010fganbound} to derive their approach, the $f$-GAN. 
\cite{uehara2016bgan} use a bound that directly follows from the density ratio estimation under Bregman divergences framework introduced by \cite{sugiyama2012bregmandre} to obtain their $b$-GAN. 
While the $b$-GAN's discriminator directly estimates the density ratio, the $f$-GAN's discriminator estimates an invertible mapping of the density ratio. 
Yet, in the case of the I-projection, both the $f$-GAN and the $b$-GAN yield the same objective, as we show in Appendix \ref{ap:GAN}. 
For both the $f$-GAN and $b$-GAN the desired $f$-divergence determines the discriminator objective.
\cite{uehara2016bgan} note that the discriminator objective, implied by the I-projection, is unstable.
As both approaches are formulated in a general way to minimize any $f$-divergence,  they do not exploit the special structure of the I-projection. 
Exploiting this structure permits us to apply a tight upper bound of the I-projection for latent variable models, which results in a higher quality of the estimated models.

\cite{li2019adversarial} introduce an adversarial approach to compute the I-projection based on density ratios, estimated by logistic regression. 
Yet, their approach assumes access to the unnormalized target density, i.e., they are working in a variational inference setting.
The most important difference to GANs is that we do not base EIM on an adversarial formulation and no adversarial game has to be solved.
This removes a major source of instability in the training process, which we discuss in more detail in Section \ref{sec:gan_comp}.
\section{Expected Information Maximization}
\label{sec:EIM}
Expected Information Maximization (EIM) is a general algorithm for minimizing the I-projection for any latent variable model. 
We first derive EIM for general marginal latent variable models, i.e., $q(x) = \int q(x|z)q(z)dz$ 
and subsequently extend our derivations to conditional latent variable models, i.e.,
$q(x|y) = \int q(x|z,y)q(z|y)dz.$
EIM uses an upper bound for the objective of the marginal distribution.
Similar to Expectation-Maximization (EM), our algorithm iterates between an M-step and an E-step.
In the corresponding M-step, we minimize the upper bound and in the E-step we tighten it using a variational distribution.

\subsection{EIM for Latent Variable Models}
The I-projection can be simplified using a (tight) variational upper bound \citep{arenz2018vips} which can be obtained by introducing an auxiliary distribution $\tilde{q}(z|x)$ and using Bayes rule
\begin{align*}
&\KL{q(x)}{p(x)} = \underbrace{U_{\tilde{q}, p}(q)}_\textrm{upper bound} - \underbrace{\mathbb{E}_{q(x)}[ \KL {q(z|x)}{\tilde{q}(z|x)}}_{\geq 0}],
\end{align*}
where
\begin{align}U_{\tilde{q}, p}(q) = \iint q(x|z) q(z)  \left(\log \dfrac{q(x|z)q(z)}{p(x)} - \log \tilde{q}(z|x) \right) dzdx.\label{eq:upperbound}\end{align}
The derivation of the bound is given in Appendix \ref{ap:bound}. 
It is easy to see that $U_{\tilde{q}, p}(q)$ is an upper bound as the expected KL term is always non-negative. 
In the corresponding E-step, the model from the previous iteration, which we denote as $\old{q}(x)$, is used to tighten the bound by setting 
$\tilde{q}(z|x) = \old{q}(x|z)\old{q}(z) / \old{q}(x).$
In the M-step, we update the model distribution by minimizing the upper bound $U_{\tilde{q}, p}(q)$.
Yet, opposed to \cite{arenz2018vips}, we cannot work directly with the upper bound since it still depends on $\log p(x)$, which we cannot evaluate. 
However, we can reformulate the upper bound by setting the given relation for $\tilde{q}(z|x)$ of the E-step into Eq.~\ref{eq:upperbound},  

\begin{align}U_{\old{q}, p}(q) =
 \int q(z) \left( \int q(x|z) \log \dfrac{\old{q}(x)}{p(x)} dx  +  \KL{q(x | z)}{\old{q}(x | z)} \right) dz +\KL{q(z)}{\old{q}(z)}. \label{eq:m-step}
\end{align}
The upper bound now contains a density ratio between the old model distribution and the data.
This density ratio can be estimated using samples of $\old{q}$ and $p$, for example, by using logistic regression as shown in Section \ref{sec:dre}. 
We can use the logits $\phi(x)$ of such a logistic regressor to estimate the log density ratio $\log (\old{q}(x) / p(x) )$ in Equation \ref{eq:m-step}. 
This yields an upper bound $U_{ \old{q}, \phi}(q)$ that depends on $\phi(x)$ instead of $p(x)$. 
Optimizing this bound corresponds to the M-step of our approach. 
In the E-step, we set $\old{q}$ to the newly obtained $q$ and retrain the density ratio estimator $\phi(x)$.
Both steps formally result in the following bilevel optimization problem
\begin{align*}
    q_{t+1} \in \textrm{argmin}_{q(x)} U_{ \old{q}, \phi^*}(q) \quad \textrm{s.t.} \quad \phi^*(x) \in \textrm{argmin}_{\phi(x)} \textrm{BCE}(\phi(x), p(x), \old{q}(x)) .
\end{align*}
Using a discriminator also comes with the advantage that we can use additional discriminative features $g(x)$ as input to our discriminator that are not directly available for the generator.
For example, if $x$ models trajectories of pedestrians, $g(x)$ could indicate whether the trajectory reaches any positions that are not plausible such as rooftops or trees.
These features simplify the discrimination task and can therefore improve our model accuracy which is not possible with M-projection based algorithms such as EM.

\subsection{EIM for Conditional Latent Variable Models}

For conditional distributions, we aim at finding the conditional I-projection
$$
\textrm{argmin}_{q(x|y)} \mathbb{E}_{p(y)}\left[ \KL{q(x|y)}{p(x|y)}\right].$$

The derivations for the conditional upper bound follow the same steps as the derivations in the marginal case, where all distributions are extended by the context variable $y$. 
We refer to the supplement for details. 
The log density ratio estimator $\phi(x, y)$ now discriminates between samples of the joint distribution of $x$ and $y$.
For training $\phi( x, y)$ we generate a new sample $x$ for each context $y$,  using the distribution $q_\textrm{old}(x|y)$.
Hence, as the context distribution is the same for the true data and the generated data, the log density ratio of the conditional distributions is equal to the log density ratio of the joint distributions. 

\subsection{Relation to GANs and EM}
\label{sec:gan_comp}
\begin{figure}
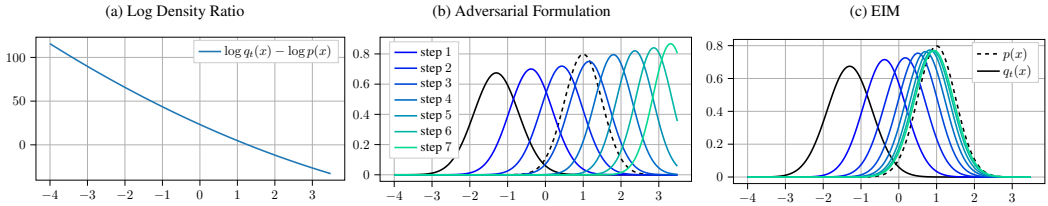

\centering
\begin{minipage}[t]{0.33\textwidth}
\subcaption*{\tiny{(a) Log Density Ratio}}
\resizebox{\textwidth}{!}{\input{img/kl_effect/ldr.tex}}
\end{minipage}%
\begin{minipage}[t]{0.33\textwidth}
\subcaption*{\tiny{(b) Adversarial Formulation}}
\resizebox{\textwidth}{!}{\input{img/kl_effect/without_kl.tex}}
\end{minipage}
\begin{minipage}[t]{0.33\textwidth}
\subcaption*{\tiny{(c) EIM}}
\resizebox{\textwidth}{!}{\input{img/kl_effect/with_kl.tex}}
\end{minipage}%
\caption{An illustrative example of the benefits of EIM versus an adversarial formulation.
(a): The true log density ratio of the model and the target distribution (both are Gaussians).
The location of the optimum is unbounded. 
(b): In the adversarial formulation, the generator minimizes the expected log density ratio.
If we neglect that the adversarial discriminator changes with every update step of the generator, the generator updates yield an unbounded solution. 
Hence, too aggressive updates of the generator yield unstable behavior.
(c): The upper bound of EIM introduces an additional KL-term as objective.
Optimizing this objective directly yields the optimal solution without the need to recompute the density ratio estimate.\label{fig:KL}} 
\end{figure}

There is a close relation of EIM to GANs due to the use of a logistic discriminator for the density ratio estimation. 
It is therefore informative to investigate the differences in the case without latent variables. 
In an adversarial formulation, the density ratio estimator would directly replace the density ratio in the original I-projection equation, i.e.,  
\begin{align*}
    \textrm{argmin}_{q(x)} \int q(x) \phi^*(x) dx \quad \textrm{s.t.} \quad \phi^*(x) \in \textrm{argmin}_{\phi(x)} \textrm{BCE}(\phi(x), p(x), q^*(x)).
\end{align*}
However, such adversarial games are often hard to optimize.
In contrast, EIM offers a bilevel optimization problem where the discriminator is explicitly learned on the old data distribution $\old{q}(x)$,  
\begin{align*}
    \textrm{argmin}_{q(x)} \int q(x) \phi^*(x) dx + \KL{q(x)}{\old{q}(x)}  \text{ } \textrm{s.t.} \text{ } \phi^*(x) \in  \textrm{argmin}_{\phi(x)} \textrm{BCE}(\phi(x), p(x), \old{q}(x)).
\end{align*}
Thus, there is no circular dependency between the optimal generator and the optimal discriminator.
\autoref{fig:KL} illustrates that the proposed non-adversarial formulation does not suffer from too large model updates.
Choosing the number and step-size of the updates is thus far less critical.

EIM can also be seen as the counter-part of Expectation-Maximization (EM). 
While EM optimizes the M-projection with latent variable models, EIM uses the I-projection. However, both approaches decompose the corresponding projections into an upper bound (or lower bound for EM) and a KL-term that depends on the conditional distribution $q(z|x)$ to tighten this bound.
The exact relationship is discussed in Appendix \ref{ap:EM}.

\subsection{EIM for Gaussian Mixtures Models}
We consider Gaussian mixture models with $d$ components, i.e., multivariate Gaussian distributions $q(\cvec{x}|z_i) = \mathcal{N}(\cvec{\mu}_i, \cmat{\Sigma}_i)$ and a categorical distribution $q(z) = \textrm{Cat}(\cvec{\pi})$ for the coefficients. 
As the latent distribution $q(z)$ is discrete, the upper bound in EIM (\autoref{eq:m-step}) simplifies, as
the integral over $z$ can be written as a sum. 
Similar to the EM-algorithm, this objective can be updated individually for the coefficients and the components. 
For both updates, we will use similar update rules as defined in the VIPS algorithm \citep{arenz2018vips}.
VIPS uses a trust region optimization for the components and the coefficients, where both updates can be solved in closed form as the components are Gaussian. 
The trust regions prevent the new model from going too far away from $\old{q}$ where the density ratio estimator is inaccurate, and hence, further stabilize the learning process. 
We will now sketch both updates, where we refer to  Appendix \ref{ap:VIPS} for the full details. 

For updating the coefficients, we assume that the components have not yet been updated, and therefore $\KL{q(\cvec{x}|z_i)}{\old{q}(\cvec{x}| z_i)} = 0$ for all $z_i$. 
The objective for the coefficients thus simplifies to
\begin{align}
\textrm{argmin}_{q(z)} \sum_{i=1}^d q(z_i) \phi(z_i) +\KL{q(z)}{\old{q}(z)} \quad \text{with} \quad  \phi(z_i) = \mathbb{E}_{q(\cvec{x}|z_i)} \left[ \phi(\cvec{x}) \right],  \label{eq:EIM_weight}
\end{align}
where $\phi(z_i)$ can be approximated using samples from the corresponding component. 
This objective can easily be optimized in closed form, as shown in the VIPS algorithm \citep{arenz2018vips}.
We also use a KL trust-region to specify the step size of the update. 
For updating the individual components, the objective simplifies to
\begin{align}
\textrm{argmin}_{q(\cvec{x}|z_i)} \mathbb{E}_{q(\cvec{x}|z_i)} \left[\phi(\cvec{x})\right] + \KL{q(\cvec{x}|z_i)}{\old{q}(\cvec{x}|z_i)}. \label{eq:EIM_component}
\end{align}
As in VIPS, this optimization problem can be solved in closed form  using the MORE algorithm \citep{abdolmaleki2015more}.
The MORE algorithm uses a quadratic surrogate function that locally approximates $\phi(\cvec{x})$.
The resulting solution optimizes Equation \ref{eq:EIM_component} under a KL trust-region.
The pseudo-code of EIM for GMMs can be found in Appendix \ref{ap:code}.

\subsection{EIM for Gaussian Mixture of Experts}
In the conditional case, we consider mixtures of experts consisting of $d$ multivariate Gaussians, whose parameters depend on an input $\cvec{y}$ in a nonlinear fashion, i.e., $q(\cvec{x}|z_i, \cvec{y}) = \mathcal{N}\left(\psi_{\mu, i}(\cvec{y}), \psi_{\Sigma, i}(\cvec{y}) \right)$ and the gating is given by a neural network with softmax output.
We again decompose the resulting upper bound into individual update steps for the components and the gating.
Yet, closed-form solutions are no longer available and we need to resort to gradient-based updates. 
The objective for updating the gating is given by
\begin{align}
\textrm{argmin}_{q(z|\cvec{y})} \sum_{i=1}^d \left(\mathbb{E}_{p(\cvec{y}) q (\cvec{x}|z_i, \cvec{y})} \left[q(z_i|\cvec{y}) \phi(\cvec{x}, \cvec{y}) \right]\right) + \mathbb{E}_{p(\cvec{y})} \left[\KL{q(z|\cvec{y})}{\old{q}(z|\cvec{y})}\right]. 
\end{align}
We minimize this equation w.r.t. the parameters of the gating by gradient descent using the Adam \citep{kingma2014adam} algorithm. 
The objective for updating a single component $i$ is given by
\begin{align}
 \textrm{argmin}_{q\left(\cvec{x}|z_i, \cvec{y}\right)}
\mathbb{E}_{\tilde{p}(\cvec{y} | z)} \left[ \mathbb{E}_{q(\cvec{x}|z_i, \cvec{y})}\left[\phi(\cvec{x}, \cvec{y})\right]  + \KL{q(\cvec{x}|z_i, \cvec{y})}{\old{q}(\cvec{x}| z_i, \cvec{y})}\right],\label{eq:DNN_components}  
\end{align}

where $\tilde{p}(\cvec{y} | z) = p(\cvec{y})  q(z_i|\cvec{y}) /q(z_i)$. 
Note that we normalized the objective by $q(z_i) = \int p(\cvec{y}) q(z_i|\cvec{y})d\cvec{y}$ to insure that also components with a low prior $q(z_i)$ get large enough gradients for the updates.
As we have access to the derivatives of the density ratio estimator w.r.t. $\cvec{x}$, we can optimize Equation \ref{eq:DNN_components} with gradient descent using the reparametrization trick \citep{kingma2013vae} and Adam.

\section{Evaluation}
 We compare our approach to GANs and perform an ablation study on a toy task, with data sampled from known mixture models.
 We further apply our approach to two synthetic datasets, learning the joint configurations of a planar robot as well as a non-linear obstacle avoidance task, and two real datasets, namely the Stanford Drone Dataset \citep{Robicquet2016} and a traffic dataset from the Next Generation Simulation program.
 A full overview of all hyperparameters and network architectures can be found in Appendix \ref{ap:experiments}.
\subsection{Comparison to Generative Adversarial Approaches and Ablation Study}
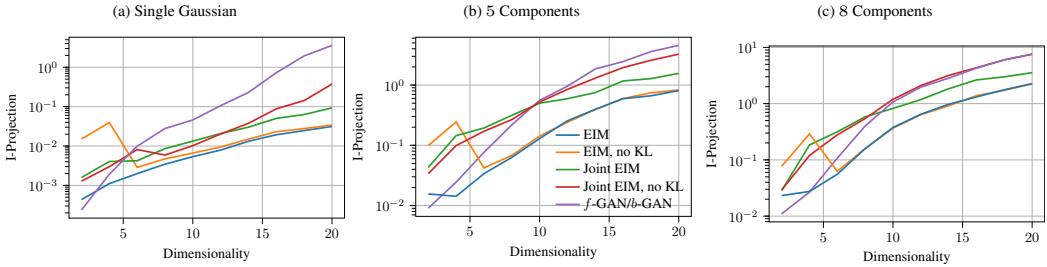
\begin{figure}[t]
\centering
\begin{minipage}[t]{0.33\textwidth}
\subcaption*{\tiny{(a) Single Gaussian}}
\resizebox{\textwidth}{!}{
\begin{tikzpicture}

\definecolor{color0}{rgb}{0.12156862745098,0.466666666666667,0.705882352941177}
\definecolor{color1}{rgb}{1,0.498039215686275,0.0549019607843137}
\definecolor{color2}{rgb}{0.172549019607843,0.627450980392157,0.172549019607843}
\definecolor{color3}{rgb}{0.83921568627451,0.152941176470588,0.156862745098039}
\definecolor{color4}{rgb}{0.580392156862745,0.403921568627451,0.741176470588235}

\begin{axis}[
log basis y={10},
tick align=outside,
tick pos=left,
x grid style={white!69.01960784313725!black},
xlabel={Dimensionality},
xmajorgrids,
xmin=1.1, xmax=20.9,
xtick style={color=black},
y grid style={white!69.01960784313725!black},
ylabel={I-Projection},
ymajorgrids,
ymin=0.00014742419535929, ymax=5.79687721045626,
ymode=log,
ytick style={color=black},
y= 0.98cm
]
\addplot [very thick, color0]
table {%
2 0.00043434292456368
4 0.00110777809168212
6 0.0019865056470735
8 0.00345408048015088
10 0.00536837282124907
12 0.00788705141749233
14 0.0131960259284824
16 0.0193104368634522
18 0.0246565479785204
20 0.0313384753651917
};
\addplot [very thick, color1]
table {%
2 0.0150706048392749
4 0.0392852830991615
6 0.00286184858996421
8 0.00471373584587127
10 0.00660638106055558
12 0.00943684158846736
14 0.0149337200447917
16 0.0230883416254073
18 0.0277572869323194
20 0.0340703211724758
};
\addplot [very thick, color2]
table {%
2 0.00158525251463288
4 0.00404029185883701
6 0.0042187217855826
8 0.00860338276252151
10 0.0133075548801571
12 0.0207771644927561
14 0.0300834595225751
16 0.050320185534656
18 0.0630826741456985
20 0.0930783178657293
};
\addplot [very thick, color3]
table {%
2 0.00127879568972276
4 0.00298975685727783
6 0.00814745756797492
8 0.00588959582382813
10 0.0102179942885414
12 0.0201546681113541
14 0.0377504213713109
16 0.0880213461816311
18 0.142825830169022
20 0.37373327370733
};
\addplot [very thick, color4]
table {%
2 0.000238459908086952
4 0.00191331227688352
6 0.00989516589324921
8 0.0277820809744298
10 0.0458778982982039
12 0.106003973633051
14 0.229575331322849
16 0.72713945582509
18 1.94340623617172
20 3.58383077979088
};
\end{axis}

\end{tikzpicture}}
\end{minipage}%
\begin{minipage}[t]{0.33\textwidth}
\subcaption*{\tiny{(b) $5$ Components}}
\resizebox{\textwidth}{!}{
\begin{tikzpicture}

\definecolor{color0}{rgb}{0.12156862745098,0.466666666666667,0.705882352941177}
\definecolor{color1}{rgb}{1,0.498039215686275,0.0549019607843137}
\definecolor{color2}{rgb}{0.172549019607843,0.627450980392157,0.172549019607843}
\definecolor{color3}{rgb}{0.83921568627451,0.152941176470588,0.156862745098039}
\definecolor{color4}{rgb}{0.580392156862745,0.403921568627451,0.741176470588235}

\begin{axis}[
legend cell align={left},
legend style={at={(1.0,0.0)}, anchor=south east, draw=none, fill=none}, 
log basis y={10},
tick align=outside,
tick pos=left,
x grid style={white!69.01960784313725!black},
xlabel={Dimensionality},
xmajorgrids,
xmin=1.1, xmax=20.9,
xtick style={color=black},
y grid style={white!69.01960784313725!black},
ylabel={I-Projection},
ymajorgrids,
ymin=0.00662010337872964, ymax=6.21729194271521,
ymode=log,
ytick style={color=black},
y=1.5cm
]

\addplot [very thick, color4, forget plot]
table {%
2 0.00903626913204789
4 0.0244903592159972
6 0.0763152560219169
8 0.218011826276779
10 0.556267035007477
12 0.956948231160641
14 1.84937835931778
16 2.45336276888847
18 3.58614046573639
20 4.55487931966782
};
\label{plot:fgan}

\addplot [very thick, color1, forget plot]
table {%
2 0.0989728851243854
4 0.2447176001966
6 0.0420886223204434
8 0.0679572699591517
10 0.142368250712752
12 0.241513660550117
14 0.39238171055913
16 0.596657277643681
18 0.743351332843304
20 0.828832769393921
};
\label{plot:eim_no_kl}

\addplot [very thick, color2, forget plot]
table {%
2 0.0429521825630218
4 0.144932241458446
6 0.193427699431777
8 0.314485896378756
10 0.499168458580971
12 0.593395437300205
14 0.749497377872467
16 1.16936509860189
18 1.27819922268391
20 1.5625737041235
};
\label{plot:eim_joint}

\addplot [very thick, color3, forget plot]
table {%
2 0.0338862939272076
4 0.0987232824787498
6 0.170906373858452
8 0.266706990450621
10 0.522025325149298
12 0.84775198996067
14 1.3013336122036
16 1.94926553368568
18 2.57368429303169
20 3.25708997249603
};
\label{plot:eim_joint_no_kl}

\addplot [very thick, color0, forget plot]
table {%
2 0.0156194393290207
4 0.0142371075227857
6 0.0337806624360383
8 0.0629856992512941
10 0.129414054378867
12 0.255003146082163
14 0.394934785366058
16 0.594104175269604
18 0.663259840011597
20 0.80921433866024
};
\label{plot:eim}

\addlegendimage{very thick, color0}  \addlegendentry{EIM}
\addlegendimage{very thick, color1}  \addlegendentry{EIM, no KL}
\addlegendimage{very thick, color2}  \addlegendentry{Joint EIM}
\addlegendimage{very thick, color3}  \addlegendentry{Joint EIM, no KL}
\addlegendimage{very thick, color4}  \addlegendentry{$f$-GAN/$b$-GAN}
\end{axis}

\end{tikzpicture}}
\end{minipage}
\begin{minipage}[t]{0.33\textwidth}
\subcaption*{\tiny{(c) $8$ Components}}
\resizebox{\textwidth}{!}{
\begin{tikzpicture}

\definecolor{color0}{rgb}{0.12156862745098,0.466666666666667,0.705882352941177}
\definecolor{color1}{rgb}{1,0.498039215686275,0.0549019607843137}
\definecolor{color2}{rgb}{0.172549019607843,0.627450980392157,0.172549019607843}
\definecolor{color3}{rgb}{0.83921568627451,0.152941176470588,0.156862745098039}
\definecolor{color4}{rgb}{0.580392156862745,0.403921568627451,0.741176470588235}

\begin{axis}[
log basis y={10},
tick align=outside,
tick pos=left,
x grid style={white!69.01960784313725!black},
xlabel={Dimensionality},
xmajorgrids,
xmin=1.1, xmax=20.9,
xtick style={color=black},
y grid style={white!69.01960784313725!black},
ylabel={I-Projection},
ymajorgrids,
ymin=0.00790913685521238, ymax=10.5942921345183,
ymode=log,
ytick style={color=black},
y=1.4cm
]

\addplot [very thick, color2]
table {%
2 0.0284311844035983
4 0.184022195637226
6 0.316145583987236
8 0.582018849253654
10 0.823828187584877
12 1.18052233457565
14 1.83977281749249
16 2.64205641746521
18 2.99479791522026
20 3.55522832870483
};
\addplot [very thick, color3]
table {%
2 0.0292217204347253
4 0.119902209751308
6 0.278470020741224
8 0.542908047139645
10 1.19380827248096
12 2.09726418852806
14 3.17208244800568
16 4.34591813087463
18 5.99707984924316
20 7.55120853185654
};
\addplot [very thick, color4]
table {%
2 0.010971420397982
4 0.0267026039771736
6 0.107473861426115
8 0.400365740805864
10 1.07181781232357
12 1.96421844363213
14 2.83669710159302
16 4.27619781494141
18 6.00417814254761
20 7.63727059364319
};
\addplot [very thick, color1]
table {%
2 0.0771121416240931
4 0.288586566597223
6 0.0626649284735322
8 0.159558155015111
10 0.366532249003649
12 0.634877836704254
14 0.9169508934021
16 1.38959294557571
18 1.72465180158615
20 2.24275541901588
};
\addplot [very thick, color0]
table {%
2 0.0231678865849972
4 0.0274726025760174
6 0.0555166704580188
8 0.158184938505292
10 0.376506070792675
12 0.646795566380024
14 0.974451434612274
16 1.31528035700321
18 1.76751424670219
20 2.27019332051277
};

\end{axis}

\end{tikzpicture}}
\end{minipage}%
\caption{Average I-projection achieved for EIM, the $f$-GAN, and the modified EIM versions. 
The task is to fit a model to samples from a randomly generated GMM of different dimensions. 
Both the model and the target GMM have the same number of components.  
EIM clearly outperforms the generative adversarial approaches, especially for larger dimensions.
The ablation study shows that the separated, closed-form updates clearly yield better results.
Neglecting the KL has a big influence for lower dimensions, but is out-weighted by the error of the discriminator at higher dimensions.}
\label{fig:ablation}
\end{figure}

We compare to the $f$-GAN which is the only other method capable of minimizing the I-projection solely based on samples.
We use data sampled from randomly generated GMMs with different numbers of components and dimensionalities.
To study the influence of the previously mentioned differences of EIM to generative adversarial approaches, we also perform an ablation study. 
We compare to a version of EIM where we neglect the additional KL-term (EIM, no KL), a version were we trained all components and the coefficients jointly using gradient descent (Joint EIM), and a version where we do both (Joint EIM, no KL). 
The average I-projection achieved by the various approaches can be found in \autoref{fig:ablation}. 

\subsection{Line Reaching with Planar Robot}
\begin{figure}[t]
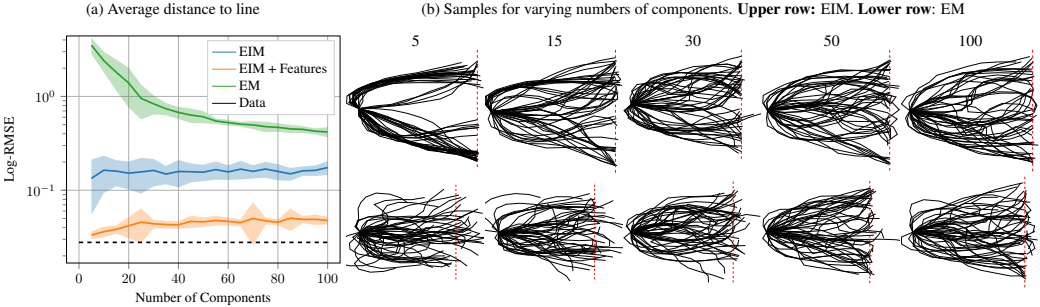

\begin{minipage}[t]{0.33\textwidth}
\subcaption*{\tiny{(a) Average distance to line}}
\resizebox{\textwidth}{!}{
\begin{tikzpicture}

\definecolor{color2}{rgb}{0.172549019607843,0.627450980392157,0.172549019607843}
\definecolor{color0}{rgb}{0.12156862745098,0.466666666666667,0.705882352941177}
\definecolor{color1}{rgb}{1,0.498039215686275,0.0549019607843137}

\begin{axis}[
legend cell align={left},
legend entries={{EIM},{EIM + Features},{EM},{Data}},
legend style={draw=white!80.0!black},
tick align=outside,
tick pos=left,
x grid style={lightgray!92.02614379084967!black},
xlabel={Number of Components},
xmajorgrids,
xmin=-5, xmax=105,
y grid style={lightgray!92.02614379084967!black},
ylabel={Log-RMSE},
ymajorgrids,
ymin=-0.183265932357131, ymax=4.41137503830352,
ymode=log
]
\addlegendimage{no markers, color0}
\addlegendimage{no markers, color1}
\addlegendimage{no markers, color2}
\addlegendimage{no markers, black}
\path [fill=color0, fill opacity=0.3] (axis cs:5,0.213299122844378)
--(axis cs:5,0.0546656354416948)
--(axis cs:10,0.0936057307450446)
--(axis cs:15,0.110347456281645)
--(axis cs:20,0.101838066018646)
--(axis cs:25,0.0921331497088715)
--(axis cs:30,0.109874027064918)
--(axis cs:35,0.115976027951161)
--(axis cs:40,0.107090365381873)
--(axis cs:45,0.121483363714529)
--(axis cs:50,0.125186439493498)
--(axis cs:55,0.129465297086804)
--(axis cs:60,0.128843658268879)
--(axis cs:65,0.130376836050913)
--(axis cs:70,0.133396385435786)
--(axis cs:75,0.13590875812517)
--(axis cs:80,0.12751026672102)
--(axis cs:85,0.125958936873235)
--(axis cs:90,0.142292529480122)
--(axis cs:95,0.142078592739436)
--(axis cs:100,0.145055922627442)
--(axis cs:100,0.20410038318901)
--(axis cs:100,0.20410038318901)
--(axis cs:95,0.183937184693531)
--(axis cs:90,0.179688300020987)
--(axis cs:85,0.17322912104034)
--(axis cs:80,0.190059170861667)
--(axis cs:75,0.201059033250762)
--(axis cs:70,0.184479651096101)
--(axis cs:65,0.205783269870657)
--(axis cs:60,0.185051581260635)
--(axis cs:55,0.202770808941484)
--(axis cs:50,0.186784442748753)
--(axis cs:45,0.193572175742209)
--(axis cs:40,0.209871542863782)
--(axis cs:35,0.18163321119703)
--(axis cs:30,0.215689376336472)
--(axis cs:25,0.221565419602404)
--(axis cs:20,0.202585242482843)
--(axis cs:15,0.209304480788593)
--(axis cs:10,0.233502052299715)
--(axis cs:5,0.213299122844378)
--cycle;

\path [fill=color1, fill opacity=0.3] (axis cs:5,0.0364793678745857)
--(axis cs:5,0.0300886217784726)
--(axis cs:10,0.0315317832641812)
--(axis cs:15,0.0344893664053812)
--(axis cs:25,0.0272959002168653)
--(axis cs:30,0.0391702975493179)
--(axis cs:35,0.0390906242549769)
--(axis cs:40,0.0388274341658482)
--(axis cs:45,0.0395094080235362)
--(axis cs:50,0.0405792369410302)
--(axis cs:55,0.0420300668403799)
--(axis cs:60,0.0420107138843116)
--(axis cs:65,0.0412086202947038)
--(axis cs:70,0.0255813844910801)
--(axis cs:75,0.0430815223374511)
--(axis cs:80,0.0421343742514109)
--(axis cs:85,0.036811726315213)
--(axis cs:90,0.0445633876512367)
--(axis cs:95,0.0426793225406234)
--(axis cs:100,0.0431509259032072)
--(axis cs:100,0.0522510803480784)
--(axis cs:100,0.0522510803480784)
--(axis cs:95,0.0552496593934229)
--(axis cs:90,0.0532202473207309)
--(axis cs:85,0.0639363518044815)
--(axis cs:80,0.0490668326057175)
--(axis cs:75,0.0513398818075019)
--(axis cs:70,0.0744551347700323)
--(axis cs:65,0.0492032726976425)
--(axis cs:60,0.0511883447910565)
--(axis cs:55,0.0533315856729105)
--(axis cs:50,0.0514006919028678)
--(axis cs:45,0.0536478268866783)
--(axis cs:40,0.0468838851601414)
--(axis cs:35,0.0471235636496392)
--(axis cs:30,0.0487750391047735)
--(axis cs:25,0.0639503215995259)
--(axis cs:15,0.0422419849256738)
--(axis cs:10,0.0409220718162553)
--(axis cs:5,0.0364793678745857)
--cycle;

\path [fill=color2, fill opacity=0.3] (axis cs:5,4.20252772145531)
--(axis cs:5,2.83309784978892)
--(axis cs:10,1.80464694343838)
--(axis cs:15,1.14637702513762)
--(axis cs:20,0.759901608160842)
--(axis cs:25,0.568324066625503)
--(axis cs:30,0.641508126286749)
--(axis cs:35,0.628642909092574)
--(axis cs:40,0.575887966323001)
--(axis cs:45,0.520702700381955)
--(axis cs:50,0.517505946617813)
--(axis cs:55,0.509117591235909)
--(axis cs:60,0.489099166536323)
--(axis cs:65,0.476827978838167)
--(axis cs:70,0.44770707460765)
--(axis cs:75,0.415604167458782)
--(axis cs:80,0.419912089770555)
--(axis cs:85,0.404273289008341)
--(axis cs:90,0.407824527632335)
--(axis cs:95,0.394592345820303)
--(axis cs:100,0.36625309078938)
--(axis cs:100,0.469648247165586)
--(axis cs:100,0.469648247165586)
--(axis cs:95,0.45725249555151)
--(axis cs:90,0.482879428861161)
--(axis cs:85,0.498156631598034)
--(axis cs:80,0.516588258569478)
--(axis cs:75,0.537442678519617)
--(axis cs:70,0.54552801059151)
--(axis cs:65,0.536968685919314)
--(axis cs:60,0.562655772745677)
--(axis cs:55,0.586084850592186)
--(axis cs:50,0.696551339542195)
--(axis cs:45,0.744600595512797)
--(axis cs:40,0.774579424478876)
--(axis cs:35,0.850936951357103)
--(axis cs:30,1.02727404208782)
--(axis cs:25,1.33840540163734)
--(axis cs:20,2.02069989129973)
--(axis cs:15,2.4713910003544)
--(axis cs:10,2.99951932957063)
--(axis cs:5,4.20252772145531)
--cycle;

\addplot [very thick, color0]
table [row sep=\\]{%
5	0.133982379143036 \\
10	0.16355389152238 \\
15	0.159825968535119 \\
20	0.152211654250744 \\
25	0.156849284655638 \\
30	0.162781701700695 \\
35	0.148804619574095 \\
40	0.158480954122827 \\
45	0.157527769728369 \\
50	0.155985441121125 \\
55	0.166118053014144 \\
60	0.156947619764757 \\
65	0.168080052960785 \\
70	0.158938018265944 \\
75	0.168483895687966 \\
80	0.158784718791344 \\
85	0.149594028956787 \\
90	0.160990414750555 \\
95	0.163007888716484 \\
100	0.174578152908226 \\
};
\addplot [very thick, color1]
table [row sep=\\]{%
5	0.0332839948265292 \\
10	0.0362269275402183 \\
15	0.0383656756655275 \\
25	0.0456231109081956 \\
30	0.0439726683270457 \\
35	0.043107093952308 \\
40	0.0428556596629948 \\
45	0.0465786174551073 \\
50	0.045989964421949 \\
55	0.0476808262566452 \\
60	0.046599529337684 \\
65	0.0452059464961732 \\
70	0.0500182596305562 \\
75	0.0472107020724765 \\
80	0.0456006034285642 \\
85	0.0503740390598472 \\
90	0.0488918174859838 \\
95	0.0489644909670231 \\
100	0.0477010031256428 \\
};
\addplot [very thick, color2]
table [row sep=\\]{%
5	3.51781278562211 \\
10	2.40208313650451 \\
15	1.80888401274601 \\
20	1.39030074973028 \\
25	0.953364734131422 \\
30	0.834391084187284 \\
35	0.739789930224839 \\
40	0.675233695400939 \\
45	0.632651647947376 \\
50	0.607028643080004 \\
55	0.547601220914047 \\
60	0.525877469641 \\
65	0.50689833237874 \\
70	0.49661754259958 \\
75	0.4765234229892 \\
80	0.468250174170017 \\
85	0.451214960303188 \\
90	0.445351978246748 \\
95	0.425922420685906 \\
100	0.417950668977483 \\
};
\addplot [very thick, black, dashed]
table [row sep=\\]{%
0	0.0278307472315945 \\
100	0.0278307472315945 \\
};
\end{axis}

\end{tikzpicture}}
\end{minipage}%
\begin{minipage}[t]{0.66\textwidth}
\subcaption*{\tiny{(b) Samples for varying numbers of components. \textbf{Upper row:} EIM. \textbf{Lower row}: EM}}
\begin{minipage}[t]{0.20\textwidth}
\centering \tiny{5} \\
\resizebox{\textwidth}{!}{\input{img/robot_line/robot_line_samples_mg5.tex}} 
\end{minipage}%
\begin{minipage}[t]{0.20\textwidth}
\centering \tiny{15} \\
\resizebox{\textwidth}{!}{\input{img/robot_line/robot_line_samples_mg15.tex}}
\end{minipage}%
\begin{minipage}[t]{0.20\textwidth}
\centering \tiny{30} \\
\resizebox{\textwidth}{!}{\input{img/robot_line/robot_line_samples_mg30.tex}}
\end{minipage}%
\begin{minipage}[t]{0.20\textwidth}
\centering \tiny{50}\\
\resizebox{\textwidth}{!}{\input{img/robot_line/robot_line_samples_mg50.tex}}
\end{minipage}%
\begin{minipage}[t]{0.20\textwidth}
\centering \tiny{100}\\
\resizebox{\textwidth}{!}{\input{img/robot_line/robot_line_samples_mg100.tex}}
\end{minipage}
\begin{minipage}{0.20\textwidth}
\resizebox{\textwidth}{!}{\input{img/robot_line/robot_line_samples_em5.tex}}
\end{minipage}%
\begin{minipage}{0.20\textwidth}
\resizebox{\textwidth}{!}{\input{img/robot_line/robot_line_samples_em15.tex}}
\end{minipage}%
\begin{minipage}{0.20\textwidth}
\resizebox{\textwidth}{!}{\input{img/robot_line/robot_line_samples_em30.tex}}
\end{minipage}%
\begin{minipage}{0.20\textwidth}
\resizebox{\textwidth}{!}{\input{img/robot_line/robot_line_samples_em50.tex}}
\end{minipage}%
\begin{minipage}{0.20\textwidth}
\resizebox{\textwidth}{!}{\input{img/robot_line/robot_line_samples_em100.tex}}
\end{minipage}
\end{minipage}
\caption{Average distance to line and samples for robot line reaching.
While EIM for small numbers of components ignores modes, not considering the whole line, it learns models that achieve the underlying task, i.e., reach the line.
Providing additional information to the density ratio estimator further decreases the average distance to the line.
EM, on the other hand, averages over the modes, and thus, fails to reach the line even for large numbers of components.}
\label{fig:robot_gmm}
\end{figure}

We extended the introductory example of the planar reaching task and collected expert data from a $10$-link planar robot tasked with reaching a point on a line.
We fitted GMMs with an increasing number of components using EIM, EIM with additional features, where the end-effector coordinates for a given joint configuration were provided, and EM. 
Even for a large number of components, we see effects similar to the introductory example, i.e., the M-projection solution provided by EM fails to reach the line while EIM manages to do so.  
For small numbers, EIM ignores parts of the line, while more and more parts of it get covered as we increase the number of components. 
With the additional features, the imitation of the line reaching was even more accurate.
See \autoref{fig:robot_gmm}, for the average distance between the end-effector and the line as well as samples from both EM and EIM. 
\subsection{Pedestrian and Traffic Prediction}
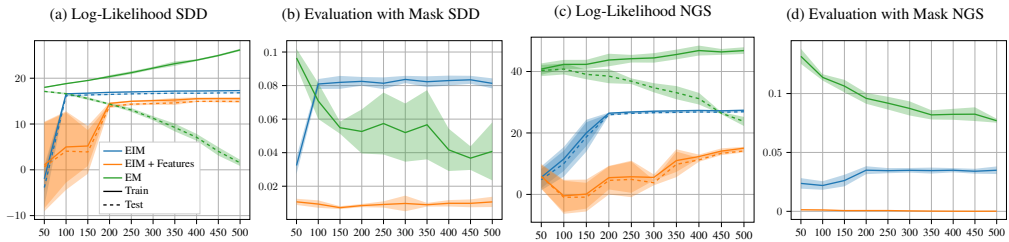
\begin{figure}[t]
    \centering
    \begin{minipage}{0.24\textwidth}
    \subcaption*{\tiny{(a) Log-Likelihood SDD}}
    \resizebox{\textwidth}{!}{
\begin{tikzpicture}

\definecolor{color0}{rgb}{0.12156862745098,0.466666666666667,0.705882352941177}
\definecolor{color1}{rgb}{1,0.498039215686275,0.0549019607843137}
\definecolor{color2}{rgb}{0.172549019607843,0.627450980392157,0.172549019607843}

\begin{axis}[
legend cell align={left},
legend style={at={(0.3,0.03)}, anchor=south west, draw=white!80.0!black},
tick align=outside,
tick pos=left,
x grid style={white!69.01960784313725!black},
xmajorgrids,
xmin=-0.45, xmax=9.45,
xtick style={color=black},
xtick={0,1,2,3,4,5,6,7,8,9},
xticklabels={50,100,150,200,250,300,350,400,450,500},
y grid style={white!69.01960784313725!black},
ymajorgrids,
ymin=-10.8374101641839, ymax=28.1995145861668,
ytick style={color=black}
]

\addlegendimage{color0, very thick} 
\addlegendentry{EIM}
\addlegendimage{color1, very thick} 
\addlegendentry{EIM + Features}
\addlegendimage{color2, very thick} 
\addlegendentry{EM}
\addlegendimage{black, very thick} 
\addlegendentry{Train}
\addlegendimage{black, dashed, very thick} 
\addlegendentry{Test}

\path [fill=color0, fill opacity=0.3]
(axis cs:0,0.981579065322876)
--(axis cs:0,-4.91379809379578)
--(axis cs:1,16.5177826732397)
--(axis cs:2,16.6609222069383)
--(axis cs:3,16.8716175109148)
--(axis cs:4,16.9221879169345)
--(axis cs:5,17.0355099402368)
--(axis cs:6,17.1547343507409)
--(axis cs:7,17.1631977781653)
--(axis cs:8,17.175426222384)
--(axis cs:9,17.2333861999214)
--(axis cs:9,17.3583839721978)
--(axis cs:9,17.3583839721978)
--(axis cs:8,17.3203623518348)
--(axis cs:7,17.2505061402917)
--(axis cs:6,17.2246067747474)
--(axis cs:5,17.1546717919409)
--(axis cs:4,17.0985373333097)
--(axis cs:3,16.9676845520735)
--(axis cs:2,16.8084534034133)
--(axis cs:1,16.6716246753931)
--(axis cs:0,0.981579065322876)
--cycle;

\path [fill=color0, fill opacity=0.3]
(axis cs:0,0.267143726348877)
--(axis cs:0,-7.99110174179077)
--(axis cs:1,16.0240539461374)
--(axis cs:2,16.2240578085184)
--(axis cs:3,16.3278790712357)
--(axis cs:4,16.493666395545)
--(axis cs:5,16.5566566288471)
--(axis cs:6,16.6149130910635)
--(axis cs:7,16.6000683307648)
--(axis cs:8,16.7219063974917)
--(axis cs:9,16.6992117539048)
--(axis cs:9,16.9276727065444)
--(axis cs:9,16.9276727065444)
--(axis cs:8,16.8423712514341)
--(axis cs:7,16.8279025554657)
--(axis cs:6,16.8553246408701)
--(axis cs:5,16.7440864741802)
--(axis cs:4,16.6681721359491)
--(axis cs:3,16.5995653867722)
--(axis cs:2,16.4593223184347)
--(axis cs:1,16.3141532987356)
--(axis cs:0,0.267143726348877)
--cycle;

\path [fill=color1, fill opacity=0.3]
(axis cs:0,10.4978324174881)
--(axis cs:0,-8.40543448925018)
--(axis cs:1,-2.79216194152832)
--(axis cs:2,0.829080581665039)
--(axis cs:3,14.2732571959496)
--(axis cs:4,14.7888788878918)
--(axis cs:5,14.8876587003469)
--(axis cs:6,14.7973993718624)
--(axis cs:7,15.2569555342197)
--(axis cs:8,15.2949892878532)
--(axis cs:9,15.2327334284782)
--(axis cs:9,15.8629181981087)
--(axis cs:9,15.8629181981087)
--(axis cs:8,15.8122189640999)
--(axis cs:7,15.8027063310146)
--(axis cs:6,15.6881581842899)
--(axis cs:5,15.3262123018503)
--(axis cs:4,15.133179217577)
--(axis cs:3,14.7251062989235)
--(axis cs:2,9.56777000427246)
--(axis cs:1,12.7765645980835)
--(axis cs:0,10.4978324174881)
--cycle;

\path [fill=color1, fill opacity=0.3]
(axis cs:0,10.3476152420044)
--(axis cs:0,-9.06300449371338)
--(axis cs:1,-4.37689828872681)
--(axis cs:2,-0.860041856765747)
--(axis cs:3,13.4711948335171)
--(axis cs:4,14.04998883605)
--(axis cs:5,14.1808963418007)
--(axis cs:6,14.1601532101631)
--(axis cs:7,14.801944449544)
--(axis cs:8,14.7234397530556)
--(axis cs:9,14.6841325461864)
--(axis cs:9,15.1553201973438)
--(axis cs:9,15.1553201973438)
--(axis cs:8,15.1938480734825)
--(axis cs:7,15.0750143975019)
--(axis cs:6,15.063775241375)
--(axis cs:5,14.7588867545128)
--(axis cs:4,14.5945328772068)
--(axis cs:3,14.2607502043247)
--(axis cs:2,8.66238951683044)
--(axis cs:1,12.5167880058289)
--(axis cs:0,10.3476152420044)
--cycle;

\path [fill=color2, fill opacity=0.3]
(axis cs:0,18.1021343893855)
--(axis cs:0,17.8863358644903)
--(axis cs:1,18.6921992504874)
--(axis cs:2,19.449056350496)
--(axis cs:3,20.0446033534728)
--(axis cs:4,20.8940039037946)
--(axis cs:5,21.9109971657664)
--(axis cs:6,22.6836314842147)
--(axis cs:7,23.7393343904212)
--(axis cs:8,24.7393546861382)
--(axis cs:9,25.9286977431786)
--(axis cs:9,26.4251089156963)
--(axis cs:9,26.4251089156963)
--(axis cs:8,25.1657405019953)
--(axis cs:7,24.1492648245441)
--(axis cs:6,23.6757105176699)
--(axis cs:5,22.5024110295132)
--(axis cs:4,21.5306972059375)
--(axis cs:3,20.6780083947858)
--(axis cs:2,19.5539856442244)
--(axis cs:1,18.9407602947978)
--(axis cs:0,18.1021343893855)
--cycle;

\path [fill=color2, fill opacity=0.3]
(axis cs:0,17.2734288283547)
--(axis cs:0,16.9702940505607)
--(axis cs:1,16.4198684144693)
--(axis cs:2,15.2596628435511)
--(axis cs:3,14.1185142121409)
--(axis cs:4,12.6463340436025)
--(axis cs:5,10.8034865202045)
--(axis cs:6,8.38559134029492)
--(axis cs:7,6.35222707277882)
--(axis cs:8,3.10815651431165)
--(axis cs:9,0.844699056432993)
--(axis cs:9,2.19576748083425)
--(axis cs:9,2.19576748083425)
--(axis cs:8,4.95572651363598)
--(axis cs:7,7.84742702811032)
--(axis cs:6,10.1011755142245)
--(axis cs:5,11.7179883331834)
--(axis cs:4,13.4876117534882)
--(axis cs:3,14.5664793276481)
--(axis cs:2,15.8577786724589)
--(axis cs:1,16.8166712069051)
--(axis cs:0,17.2734288283547)
--cycle;

\addplot [very thick, color0, forget plot]
table {%
0 -1.96610951423645
1 16.5947036743164
2 16.7346878051758
3 16.9196510314941
4 17.0103626251221
5 17.0950908660889
6 17.1896705627441
7 17.2068519592285
8 17.2478942871094
9 17.2958850860596
};

\addplot [very thick, color0, dashed, forget plot]
table {%
0 -3.86197900772095
1 16.1691036224365
2 16.3416900634766
3 16.4637222290039
4 16.5809192657471
5 16.6503715515137
6 16.7351188659668
7 16.7139854431152
8 16.7821388244629
9 16.8134422302246
};
\addplot [very thick, color1, forget plot]
table {%
0 1.04619896411896
1 4.99220132827759
2 5.19842529296875
3 14.4991817474365
4 14.9610290527344
5 15.1069355010986
6 15.2427787780762
7 15.5298309326172
8 15.5536041259766
9 15.5478258132935
};
\addplot [very thick, color1, dashed, forget plot]
table {%
0 0.642305374145508
1 4.06994485855103
2 3.90117383003235
3 13.8659725189209
4 14.3222608566284
5 14.4698915481567
6 14.611964225769
7 14.9384794235229
8 14.958643913269
9 14.9197263717651
};
\addplot [very thick, color2, forget plot]
table {%
0 17.9942351269379
1 18.8164797726426
2 19.5015209973602
3 20.3613058741293
4 21.2123505548661
5 22.2067040976398
6 23.1796710009423
7 23.9442996074826
8 24.9525475940667
9 26.1769033294375
};
\addplot [very thick, color2, dashed, forget plot]
table {%
0 17.1218614394577
1 16.6182698106872
2 15.558720758005
3 14.3424967698945
4 13.0669728985453
5 11.260737426694
6 9.24338342725973
7 7.09982705044457
8 4.03194151397382
9 1.52023326863362
};
\end{axis}

\end{tikzpicture}}
    \end{minipage}%
    \begin{minipage}{0.24\textwidth}
    \subcaption*{\tiny{(b) Evaluation with Mask SDD}}
    \resizebox{\textwidth}{!}{
\begin{tikzpicture}

\definecolor{color0}{rgb}{0.12156862745098,0.466666666666667,0.705882352941177}
\definecolor{color1}{rgb}{1,0.498039215686275,0.0549019607843137}
\definecolor{color2}{rgb}{0.172549019607843,0.627450980392157,0.172549019607843}

\begin{axis}[
legend cell align={left},
legend style={draw=white!80.0!black},
tick align=outside,
tick pos=left,
x grid style={white!69.01960784313725!black},
xmajorgrids,
xmin=-0.45, xmax=9.45,
xtick style={color=black},
xtick={0,1,2,3,4,5,6,7,8,9},
xticklabels={50,100,150,200,250,300,350,400,450,500},
y grid style={white!69.01960784313725!black},
ymajorgrids,
yticklabel style={
        /pgf/number format/fixed,
        /pgf/number format/precision=5
},
scaled y ticks=false,
ymin=0.000199434444943052, ymax=0.106770227125232,
ytick style={color=black}
]
\path [fill=color0, fill opacity=0.3]
(axis cs:0,0.0374601603178691)
--(axis cs:0,0.0272806973511653)
--(axis cs:1,0.077948819633611)
--(axis cs:2,0.0780458188957856)
--(axis cs:3,0.0798991088423409)
--(axis cs:4,0.0778801424027963)
--(axis cs:5,0.0815082919476913)
--(axis cs:6,0.0790584296905284)
--(axis cs:7,0.0800326588357745)
--(axis cs:8,0.0810425481933521)
--(axis cs:9,0.0783150306490339)
--(axis cs:9,0.0843053324133181)
--(axis cs:9,0.0843053324133181)
--(axis cs:8,0.0858293276288)
--(axis cs:7,0.0859341919929548)
--(axis cs:6,0.0855193130087613)
--(axis cs:5,0.085805572824737)
--(axis cs:4,0.0848770267632127)
--(axis cs:3,0.0851206228072251)
--(axis cs:2,0.0857637049137382)
--(axis cs:1,0.0839033245389751)
--(axis cs:0,0.0374601603178691)
--cycle;

\path [fill=color1, fill opacity=0.3]
(axis cs:0,0.0125377184240079)
--(axis cs:0,0.00884612793221416)
--(axis cs:1,0.00714121260555605)
--(axis cs:2,0.00633864765469128)
--(axis cs:3,0.00750485106835179)
--(axis cs:4,0.0072505165815793)
--(axis cs:5,0.0050435613849562)
--(axis cs:6,0.00786820599278674)
--(axis cs:7,0.00789302025560824)
--(axis cs:8,0.00745281592953582)
--(axis cs:9,0.0075844693877898)
--(axis cs:9,0.0137125577103423)
--(axis cs:9,0.0137125577103423)
--(axis cs:8,0.0121443660985854)
--(axis cs:7,0.011733393845944)
--(axis cs:6,0.0100452086057107)
--(axis cs:5,0.0143934288807633)
--(axis cs:4,0.0109973129369685)
--(axis cs:3,0.00941701159936723)
--(axis cs:2,0.00808913450789751)
--(axis cs:1,0.0115790072196362)
--(axis cs:0,0.0125377184240079)
--cycle;

\path [fill=color2, fill opacity=0.3]
(axis cs:0,0.101926100185219)
--(axis cs:0,0.0909915530639258)
--(axis cs:1,0.0600754176029088)
--(axis cs:2,0.051279434692352)
--(axis cs:3,0.0394866775166323)
--(axis cs:4,0.0388872512481644)
--(axis cs:5,0.0346140177971232)
--(axis cs:6,0.0360113852969091)
--(axis cs:7,0.0290566756759543)
--(axis cs:8,0.0297070658005886)
--(axis cs:9,0.0234838919547231)
--(axis cs:9,0.0580630693712438)
--(axis cs:9,0.0580630693712438)
--(axis cs:8,0.0437683354096193)
--(axis cs:7,0.0542687649975527)
--(axis cs:6,0.0772640685310309)
--(axis cs:5,0.0693270503428399)
--(axis cs:4,0.0758509755342612)
--(axis cs:3,0.0658014045670298)
--(axis cs:2,0.0585127252655538)
--(axis cs:1,0.0815399467748865)
--(axis cs:0,0.101926100185219)
--cycle;

\addplot [very thick, color0, forget plot]
table {%
0 0.0323704288345172
1 0.0809260720862931
2 0.0819047619047619
3 0.082509865824783
4 0.0813785845830045
5 0.0836569323862142
6 0.0822888713496448
7 0.0829834254143646
8 0.083435937911076
9 0.081310181531176
};
\addplot [very thick, color1, forget plot]
table {%
0 0.010691923178111
1 0.0093601099125961
2 0.0072138910812944
3 0.00846093133385951
4 0.00912391475927388
5 0.00971849513285977
6 0.00895670729924874
7 0.00981320705077611
8 0.00979859101406063
9 0.010648513549066
};
\addplot [very thick, color2, forget plot]
table {%
0 0.0964588266245725
1 0.0708076821888977
2 0.0548960799789529
3 0.0526440410418311
4 0.0573691133912128
5 0.0519705340699816
6 0.05663772691397
7 0.0416627203367535
8 0.0367377006051039
9 0.0407734806629834
};

\end{axis}

\end{tikzpicture}}
    \end{minipage}%
    \begin{minipage}{0.24\textwidth}
    \subcaption*{\tiny{(c) Log-Likelihood NGS}}
    \resizebox{\textwidth}{!}{
\begin{tikzpicture}

\definecolor{color0}{rgb}{0.12156862745098,0.466666666666667,0.705882352941177}
\definecolor{color1}{rgb}{1,0.498039215686275,0.0549019607843137}
\definecolor{color2}{rgb}{0.172549019607843,0.627450980392157,0.172549019607843}

\begin{axis}[
tick align=outside,
tick pos=left,
x grid style={white!69.01960784313725!black},
xmajorgrids,
xmin=-0.45, xmax=9.45,
xtick style={color=black},
xtick={0,1,2,3,4,5,6,7,8,9},
xticklabels={50,100,150,200,250,300,350,400,450,500},
y grid style={white!69.01960784313725!black},
ymajorgrids,
ymin=-9.05654751348029, ymax=51.1750588897679,
ytick style={color=black}
]
\path [fill=color0, fill opacity=0.3]
(axis cs:0,9.11523127555847)
--(axis cs:0,2.02728295326233)
--(axis cs:1,7.64867901802063)
--(axis cs:2,15.7129392623901)
--(axis cs:3,26.2317401468754)
--(axis cs:4,26.6229395866394)
--(axis cs:5,26.8556747585535)
--(axis cs:6,27.0053784996271)
--(axis cs:7,27.2151805981994)
--(axis cs:8,26.9801616966724)
--(axis cs:9,27.2098115235567)
--(axis cs:9,27.5488249510527)
--(axis cs:9,27.5488249510527)
--(axis cs:8,27.4278392493725)
--(axis cs:7,27.4365108385682)
--(axis cs:6,27.3952944129705)
--(axis cs:5,27.2791328281164)
--(axis cs:4,27.0295491218567)
--(axis cs:3,26.6102474629879)
--(axis cs:2,24.319993019104)
--(axis cs:1,15.420393705368)
--(axis cs:0,9.11523127555847)
--cycle;

\path [fill=color0, fill opacity=0.3]
(axis cs:0,8.27318835258484)
--(axis cs:0,1.32997536659241)
--(axis cs:1,5.80092716217041)
--(axis cs:2,13.9184122085571)
--(axis cs:3,25.6608783602715)
--(axis cs:4,26.1861770302057)
--(axis cs:5,26.3741728067398)
--(axis cs:6,26.4438996911049)
--(axis cs:7,26.6456791907549)
--(axis cs:8,26.5370984971523)
--(axis cs:9,26.7154373824596)
--(axis cs:9,27.0892195999622)
--(axis cs:9,27.0892195999622)
--(axis cs:8,26.9330171644688)
--(axis cs:7,27.0287821739912)
--(axis cs:6,26.98705047369)
--(axis cs:5,26.8161233663559)
--(axis cs:4,26.5687439292669)
--(axis cs:3,26.2087886929512)
--(axis cs:2,23.4775342941284)
--(axis cs:1,14.2533903121948)
--(axis cs:0,8.27318835258484)
--cycle;

\path [fill=color1, fill opacity=0.3]
(axis cs:0,9.93811655044556)
--(axis cs:0,1.72825956344604)
--(axis cs:1,-5.67135953903198)
--(axis cs:2,-4.63441741466522)
--(axis cs:3,1.60868358612061)
--(axis cs:4,0.514360427856445)
--(axis cs:5,4.56962996721268)
--(axis cs:6,7.32551217079163)
--(axis cs:7,11.6727766990662)
--(axis cs:8,13.2115016579628)
--(axis cs:9,14.585813999176)
--(axis cs:9,15.489137172699)
--(axis cs:9,15.489137172699)
--(axis cs:8,14.9928535819054)
--(axis cs:7,12.9650177955627)
--(axis cs:6,14.6976010799408)
--(axis cs:5,6.48784893751144)
--(axis cs:4,10.9812822341919)
--(axis cs:3,9.27438163757324)
--(axis cs:2,4.82613575458527)
--(axis cs:1,4.74608373641968)
--(axis cs:0,9.93811655044556)
--cycle;

\path [fill=color1, fill opacity=0.3]
(axis cs:0,9.57092714309692)
--(axis cs:0,2.01847219467163)
--(axis cs:1,-6.31874722242355)
--(axis cs:2,-5.5681859254837)
--(axis cs:3,-0.0285401344299316)
--(axis cs:4,-0.913633346557617)
--(axis cs:5,2.96544790267944)
--(axis cs:6,5.62958240509033)
--(axis cs:7,10.6105790138245)
--(axis cs:8,12.7797564268112)
--(axis cs:9,13.6586748957634)
--(axis cs:9,14.6128323674202)
--(axis cs:9,14.6128323674202)
--(axis cs:8,13.9518414735794)
--(axis cs:7,11.8649840354919)
--(axis cs:6,13.9374380111694)
--(axis cs:5,4.6352596282959)
--(axis cs:4,10.6927185058594)
--(axis cs:3,9.14702367782593)
--(axis cs:2,3.774285197258)
--(axis cs:1,4.45349341630936)
--(axis cs:0,9.57092714309692)
--cycle;

\path [fill=color2, fill opacity=0.3]
(axis cs:0,42.6451533016108)
--(axis cs:0,38.9373112274032)
--(axis cs:1,40.8399480465642)
--(axis cs:2,40.8272691451274)
--(axis cs:3,41.6915884203567)
--(axis cs:4,42.9276841744515)
--(axis cs:5,42.8792739776911)
--(axis cs:6,43.9514469236836)
--(axis cs:7,45.1965593244067)
--(axis cs:8,45.4184396265624)
--(axis cs:9,45.7121530726825)
--(axis cs:9,47.8956208442707)
--(axis cs:9,47.8956208442707)
--(axis cs:8,47.3311582445016)
--(axis cs:7,48.4372585987112)
--(axis cs:6,47.1159617725813)
--(axis cs:5,45.9691656875256)
--(axis cs:4,45.4060874383385)
--(axis cs:3,45.7891354810388)
--(axis cs:2,43.8074676607303)
--(axis cs:1,43.7871170700807)
--(axis cs:0,42.6451533016108)
--cycle;

\path [fill=color2, fill opacity=0.3]
(axis cs:0,42.1411315684834)
--(axis cs:0,38.490948186792)
--(axis cs:1,39.0364306293004)
--(axis cs:2,37.4759056411145)
--(axis cs:3,36.1823364204255)
--(axis cs:4,35.9152409222465)
--(axis cs:5,33.2194756627627)
--(axis cs:6,30.9690899777128)
--(axis cs:7,29.2594370726638)
--(axis cs:8,26.0761248885331)
--(axis cs:9,22.2650309927352)
--(axis cs:9,25.2315163109373)
--(axis cs:9,25.2315163109373)
--(axis cs:8,26.7915457928776)
--(axis cs:7,33.0563136424808)
--(axis cs:6,35.2579082572834)
--(axis cs:5,36.1990672455658)
--(axis cs:4,37.7687447717364)
--(axis cs:3,40.8209699104045)
--(axis cs:2,40.6531533016283)
--(axis cs:1,42.5593686725418)
--(axis cs:0,42.1411315684834)
--cycle;

\addplot [very thick, color0]
table {%
0 5.5712571144104
1 11.5345363616943
2 20.0164661407471
3 26.4209938049316
4 26.826244354248
5 27.067403793335
6 27.2003364562988
7 27.3258457183838
8 27.2040004730225
9 27.3793182373047
};
\addplot [very thick, color0, dashed]
table {%
0 4.80158185958862
1 10.0271587371826
2 18.6979732513428
3 25.9348335266113
4 26.3774604797363
5 26.5951480865479
6 26.7154750823975
7 26.837230682373
8 26.7350578308105
9 26.9023284912109
};
\addplot [very thick, color1]
table {%
0 5.8331880569458
1 -0.462637901306152
2 0.095859169960022
3 5.44153261184692
4 5.74782133102417
5 5.52873945236206
6 11.0115566253662
7 12.3188972473145
8 14.1021776199341
9 15.0374755859375
};
\addplot [very thick, color1, dashed]
table {%
0 5.79469966888428
1 -0.932626903057098
2 -0.896950364112854
3 4.559241771698
4 4.88954257965088
5 3.80035376548767
6 9.78351020812988
7 11.2377815246582
8 13.3657989501953
9 14.1357536315918
};
\addplot [very thick, color2]
table {%
0 40.791232264507
1 42.3135325583225
2 42.3173684029288
3 43.7403619506978
4 44.166885806395
5 44.4242198326084
6 45.5337043481324
7 46.8169089615589
8 46.374798935532
9 46.8038869584766
};
\addplot [very thick, color2, dashed]
table {%
0 40.3160398776377
1 40.7978996509211
2 39.0645294713714
3 38.501653165415
4 36.8419928469915
5 34.7092714541643
6 33.1134991174981
7 31.1578753575723
8 26.4338353407054
9 23.7482736518363
};
\end{axis}

\end{tikzpicture}}
    \end{minipage}%
    \begin{minipage}{0.24\textwidth}
    \subcaption*{\tiny{(d) Evaluation with Mask NGS}}
    \resizebox{\textwidth}{!}{
\begin{tikzpicture}

\definecolor{color0}{rgb}{0.12156862745098,0.466666666666667,0.705882352941177}
\definecolor{color1}{rgb}{1,0.498039215686275,0.0549019607843137}
\definecolor{color2}{rgb}{0.172549019607843,0.627450980392157,0.172549019607843}

\begin{axis}[
tick align=outside,
tick pos=left,
x grid style={white!69.01960784313725!black},
xmajorgrids,
xmin=-0.45, xmax=9.45,
xtick style={color=black},
xtick={0,1,2,3,4,5,6,7,8,9},
xticklabels={50,100,150,200,250,300,350,400,450,500},
y grid style={white!69.01960784313725!black},
ymajorgrids,
ymin=-0.00687755534728768, ymax=0.144897079406862,
ytick style={color=black},
yticklabel style={
        /pgf/number format/fixed,
        /pgf/number format/precision=5
},
scaled y ticks=false,
]
\path [fill=color0, fill opacity=0.3]
(axis cs:0,0.0282121598550673)
--(axis cs:0,0.0192038401449327)
--(axis cs:1,0.0182983091040696)
--(axis cs:2,0.0212122354908096)
--(axis cs:3,0.0315561068900572)
--(axis cs:4,0.0323648045773397)
--(axis cs:5,0.0328595736501965)
--(axis cs:6,0.032040094119384)
--(axis cs:7,0.0330714022959478)
--(axis cs:8,0.0319459990403073)
--(axis cs:9,0.0316501327187676)
--(axis cs:9,0.0381298672812324)
--(axis cs:9,0.0381298672812324)
--(axis cs:8,0.0361140009596927)
--(axis cs:7,0.0367125977040522)
--(axis cs:6,0.037139905880616)
--(axis cs:5,0.0368004263498035)
--(axis cs:4,0.0366991954226604)
--(axis cs:3,0.0382918931099428)
--(axis cs:2,0.0310637645091904)
--(axis cs:1,0.0255576908959304)
--(axis cs:0,0.0282121598550673)
--cycle;

\path [fill=color1, fill opacity=0.3]
(axis cs:0,0.00150009521111562)
--(axis cs:0,0.00116390478888438)
--(axis cs:1,0.000818536545880262)
--(axis cs:2,0.00025727655619489)
--(axis cs:3,0.000517359479637529)
--(axis cs:4,0.000289864889533573)
--(axis cs:5,0.000321121594224482)
--(axis cs:6,9.21539601140167e-05)
--(axis cs:7,7.9999999999969e-05)
--(axis cs:8,2.12916869918752e-05)
--(axis cs:9,9.17157287525188e-05)
--(axis cs:9,0.000148284271247499)
--(axis cs:9,0.000148284271247499)
--(axis cs:8,0.00010670831300812)
--(axis cs:7,0.000120000000000009)
--(axis cs:6,0.000299846039885976)
--(axis cs:5,0.000438878405775538)
--(axis cs:4,0.00103013511046642)
--(axis cs:3,0.000789307187029112)
--(axis cs:2,0.00101272344380516)
--(axis cs:1,0.00148813012078642)
--(axis cs:0,0.00150009521111562)
--cycle;

\path [fill=color2, fill opacity=0.3]
(axis cs:0,0.137998232372583)
--(axis cs:0,0.125173767627417)
--(axis cs:1,0.111042443567081)
--(axis cs:2,0.101024656396724)
--(axis cs:3,0.091311668414708)
--(axis cs:4,0.0864082462274814)
--(axis cs:5,0.0813484888643847)
--(axis cs:6,0.0779641792301583)
--(axis cs:7,0.0777039536283975)
--(axis cs:8,0.07664)
--(axis cs:9,0.07524)
--(axis cs:9,0.07862)
--(axis cs:9,0.07862)
--(axis cs:8,0.08832)
--(axis cs:7,0.0870200463716026)
--(axis cs:6,0.0858198207698417)
--(axis cs:5,0.0933595111356153)
--(axis cs:4,0.0972597537725186)
--(axis cs:3,0.100528331585292)
--(axis cs:2,0.111627343603276)
--(axis cs:1,0.116389556432919)
--(axis cs:0,0.137998232372583)
--cycle;

\addplot [very thick, color0]
table {%
0 0.023708
1 0.021928
2 0.026138
3 0.034924
4 0.034532
5 0.03483
6 0.03459
7 0.034892
8 0.03403
9 0.03489
};
\addplot [very thick, color1]
table {%
0 0.001332
1 0.00115333333333334
2 0.000635000000000024
3 0.000653333333333321
4 0.000659999999999994
5 0.00038000000000001
6 0.000195999999999996
7 9.9999999999989e-05
8 6.39999999999974e-05
9 0.000120000000000009
};
\addplot [very thick, color2]
table {%
0 0.131586
1 0.113716
2 0.106326
3 0.09592
4 0.091834
5 0.087354
6 0.081892
7 0.082362
8 0.08248
9 0.07693
};
\end{axis}

\end{tikzpicture}}
    \end{minipage}
    \caption{Results on the traffic prediction tasks. 
    Naturally, EM achieves the highest training log-likelihood.
    Yet, for large numbers of components, a severe amount of overfitting is observed.
    EIM, on the other hand, has no problems working with high numbers of components and achieves a higher test log-likelihood, despite optimizing a     different objective.
    We also provided a 'road mask' as additional features for the discriminator.
    We used this road mask to evaluate how realistic the generated samples are.
    While EIM without features produced more realistic samples on the Lankershim dataset, we needed the feature input to outperform EM on this evaluation on the SDD dataset. \label{fig:traffic}}
\end{figure}

We evaluated our approach on data from the Stanford Drone Dataset (SDD) \citep{Robicquet2016} and a traffic dataset from the Next Generation Simulation (NGS) program\footnote{\tiny{https://data.transportation.gov/Automobiles/Next-Generation-Simulation-NGSIM-Vehicle-Trajector/8ect-6jqj}}.  
The SDD data consists of trajectories of pedestrians, bikes, and cars and we only used the data corresponding to a single video of a single scene (Video 1, deathCircle). 
The NGS data consists of trajectories of cars where we considered the data recorded on Lankershim Boulevard. 
In both cases we extracted trajectories of length $5$, yielding highly multimodal data due to pedestrians, bikes, and cars moving at different speeds and in different directions.
We evaluated on the achieved log-likelihood of EIM and EM, see Figure \ref{fig:traffic}. 
EM achieves the highest likelihood as it directly optimizes this measure.
However, we can already see that EM massively overfits when we increase the number of components as the test-set likelihood degrades. EIM, on the other hand, produced better models with an increasing number of components. 
Additionally, we generated a mask indicating whether a given point is on the road or not and evaluated how realistic the learned models are by measuring the amount of samples violating the mask, i.e., predicting road users outside of the road.
We also evaluate a version of EIM where we provide additional features indicating if the mask is violated.
EIM achieves a much better value on this mask for the NGS dataset, while we needed the additional mask features for the discriminator on the SDD dataset to outperform EM.
Both experiments show that EIM can learn highly multi-modal density estimates that produce more realistic samples than EM. 
They further show that the models learned by EIM can be refined by additional prior knowledge provided as feature vectors.

\subsection{Obstacle Avoidance}
\begin{figure}[t]
    \centering
    \begin{minipage}[t]{0.21\textwidth}
    \subcaption*{\tiny{(a) Success Probability}}
    \resizebox{\textwidth}{!}{
\begin{tikzpicture}

\definecolor{color0}{rgb}{0.12156862745098,0.466666666666667,0.705882352941177}
\definecolor{color1}{rgb}{0.172549019607843,0.627450980392157,0.172549019607843}

\begin{axis}[
legend cell align={left},
legend style={at={(0.97,0.03)}, anchor=south east, draw=white!80.0!black},
tick align=outside,
tick pos=left,
x grid style={white!69.01960784313725!black},
xmajorgrids,
xmin=0.65, xmax=8.35,
xtick style={color=black},
xlabel={Number of Components},
y grid style={white!69.01960784313725!black},
ymajorgrids,
ymin=0.164453005312914, ymax=0.93021129229659,
ytick style={color=black}
]
\path [fill=color0, fill opacity=0.3]
(axis cs:1,0.700269310817175)
--(axis cs:1,0.562570689182825)
--(axis cs:2,0.727922318991953)
--(axis cs:3,0.748179383202989)
--(axis cs:4,0.832940384316622)
--(axis cs:5,0.838911110892108)
--(axis cs:6,0.842753278135184)
--(axis cs:7,0.843436757484464)
--(axis cs:8,0.863135902566304)
--(axis cs:8,0.895404097433696)
--(axis cs:8,0.895404097433696)
--(axis cs:7,0.887603242515537)
--(axis cs:6,0.891466721864816)
--(axis cs:5,0.884008889107892)
--(axis cs:4,0.883939615683378)
--(axis cs:3,0.880800616797011)
--(axis cs:2,0.840037681008047)
--(axis cs:1,0.700269310817175)
--cycle;

\path [fill=color1, fill opacity=0.3]
(axis cs:1,0.222239799824192)
--(axis cs:1,0.199260200175808)
--(axis cs:2,0.32158908826246)
--(axis cs:3,0.424074056648726)
--(axis cs:4,0.525052164448039)
--(axis cs:5,0.593957276625809)
--(axis cs:6,0.546093138485695)
--(axis cs:7,0.698200742301742)
--(axis cs:8,0.65951129552672)
--(axis cs:8,0.840208704473279)
--(axis cs:8,0.840208704473279)
--(axis cs:7,0.812439257698258)
--(axis cs:6,0.762786861514305)
--(axis cs:5,0.642502723374191)
--(axis cs:4,0.542467835551961)
--(axis cs:3,0.454365943351274)
--(axis cs:2,0.35219091173754)
--(axis cs:1,0.222239799824192)
--cycle;

\addplot [very thick, color0]
table {%
1 0.63142
2 0.78398
3 0.81449
4 0.85844
5 0.86146
6 0.86711
7 0.86552
8 0.87927
};
\addlegendentry{EIM}
\addplot [very thick, color1]
table {%
1 0.21075
2 0.33689
3 0.43922
4 0.53376
5 0.61823
6 0.65444
7 0.75532
8 0.74986
};
\addlegendentry{EM}
\addplot [very thick, black, dashed]
table {%
1 0.871
8 0.871
};
\addlegendentry{Data}
\end{axis}

\end{tikzpicture}}
    \end{minipage}%
    \begin{minipage}[t]{0.34\textwidth}
    \subcaption*{\tiny{(b) Samples EIM}}
    \includegraphics[width=\textwidth]{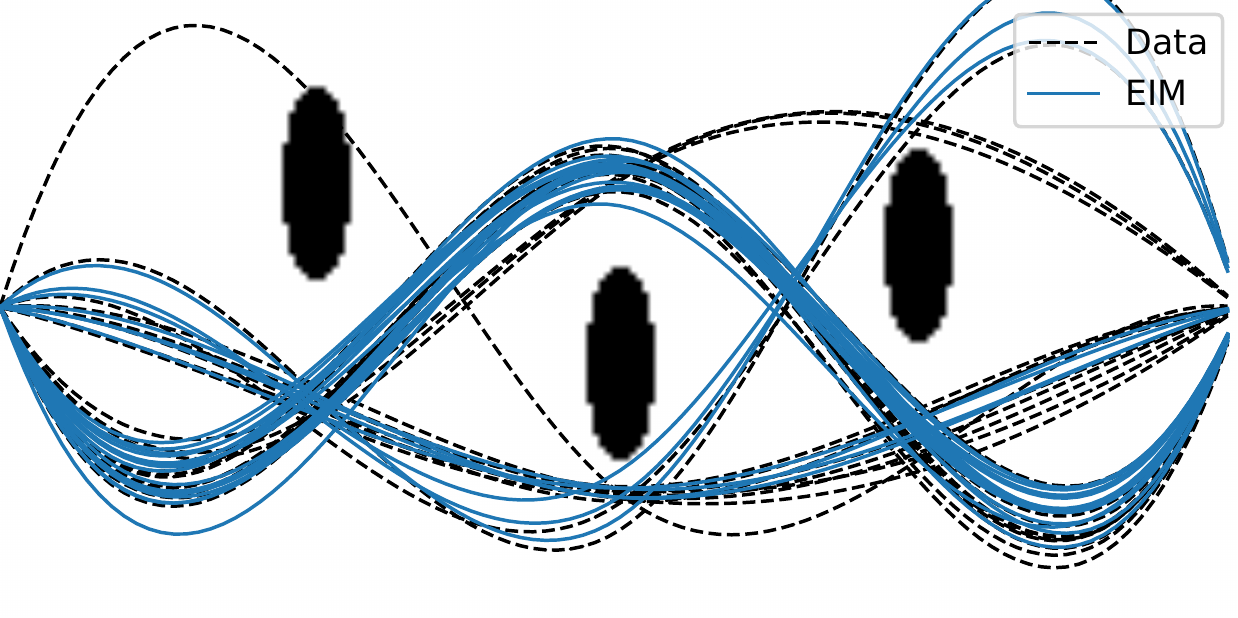}
    \end{minipage}%
    \begin{minipage}[t]{0.34\textwidth}
    \subcaption*{\tiny{(c) Samples EM}}
    \includegraphics[width=\textwidth]{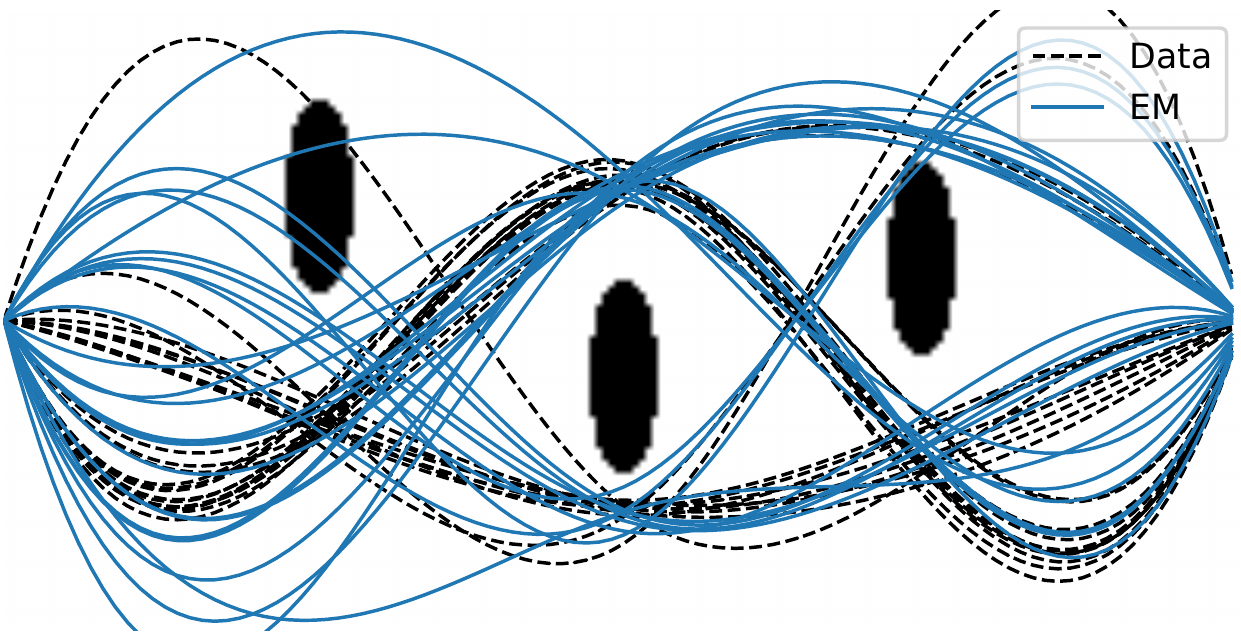}
    \end{minipage}%
    \caption{Results on the obstacle avoidance task.
    (a): Even for a small number of components, EIM has a rather high probability of success, i.e., placing a trajectory that does not hit any obstacle.
    Even with a sufficient number of components, i.e., eight, EM fails to achieve good results.
    (b) and (c): Samples of a mixture with 4 components, learned by EIM and EM respectively. 
    EM clearly averages over multiple modes in the data distribution.}
    \label{fig:obst}
\end{figure}
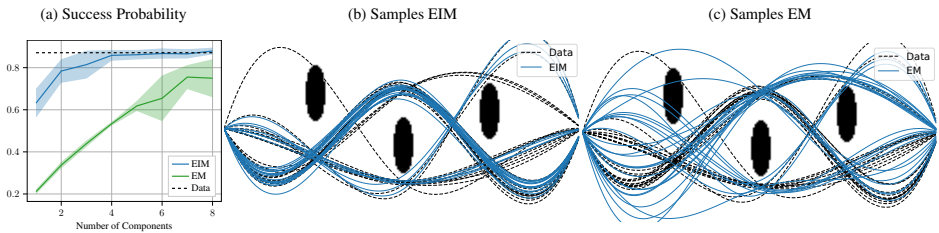
We evaluate the conditional version of EIM on an artificial obstacle avoidance task.
The context contains the location of three obstacles within an image.
The gating, as well as the components, are given by deep neural networks.
Details about the network architectures can be found in the Appendix. 
The data consists of trajectories going from the left to the right of the image. 
The trajectories are defined by setting $3$ via-points such that no obstacle is hit. 
To generate the data we sample via-points over and under the obstacles with a probability proportional to the distance between the obstacle and the image border. 
Hence, for three obstacles, there are $2^3 = 8$ different modes in the data. 
Note that, like in most real-world scenarios, the expert data is not perfect, and about $13\%$ of the trajectories in the dataset hit an obstacle. 
We fit models with various numbers of components to this data using EIM and EM and compare their performance on generating trajectories that achieve the goal. 
Results are shown in \autoref{fig:obst} together with a visualization of the task and samples produced by EIM and EM. 
EIM was able to identify most modes for the different given inputs and did not suffer from any averaging effect.
In contrast, EM does not find all modes.
As a consequence, some components of the mixture model had to average over multiple modes, resulting in poor quality trajectories.

\section{Conclusion}
We introduced Expected Information Maximization (EIM), a novel approach for computing the I-projection between general latent variable models and a target distribution, solely based on samples of the latter. 
General upper bound objectives for marginal and conditional distributions were derived, resulting in an algorithm similar to EM, but tailored for the I-projection instead of the M-projection.
We introduced efficient methods to optimize these upper bound objectives for mixture models.
In our experiments, we demonstrated the benefits of the I-projection for different behavior modelling tasks.
The introduced approach opens various pathways for future research.
While we focused on mixture models, the derived upper bounds are not exclusive to those and can be used for arbitrary latent variable models. 
Another possibility is an online adaptation of the number of used components.
\cite{arenz2018vips} propose heuristics for such an adaptation in their VIPS approach.
Those could easily be adapted to our approach.

\bibliography{main}

\begin{thebibliography}{27}
\providecommand{\natexlab}[1]{#1}
\providecommand{\url}[1]{\texttt{#1}}
\expandafter\ifx\csname urlstyle\endcsname\relax
  \providecommand{\doi}[1]{doi: #1}\else
  \providecommand{\doi}{doi: \begingroup \urlstyle{rm}\Url}\fi

\bibitem[Abdolmaleki et~al.(2015)Abdolmaleki, Lioutikov, Peters, Lau, Reis, and
  Neumann]{abdolmaleki2015more}
Abbas Abdolmaleki, Rudolf Lioutikov, Jan~R Peters, Nuno Lau, Luis~Pualo Reis,
  and Gerhard Neumann.
\newblock Model-based relative entropy stochastic search.
\newblock In \emph{Advances in Neural Information Processing Systems}, pp.\
  3537--3545, 2015.

\bibitem[Ali \& Silvey(1966)Ali and Silvey]{ali1966fdiv}
Syed~Mumtaz Ali and Samuel~D Silvey.
\newblock A general class of coefficients of divergence of one distribution
  from another.
\newblock \emph{Journal of the Royal Statistical Society. Series B
  (Methodological)}, pp.\  131--142, 1966.

\bibitem[Arenz et~al.(2018)Arenz, Zhong, and Neumann]{arenz2018vips}
O.~Arenz, M.~Zhong, and G.~Neumann.
\newblock Efficient gradient-free variational inference using policy search.
\newblock In \emph{Proceedings of the 35th International Conference on Machine
  Learning}, pp.\  234--243. pmlr, 2018.

\bibitem[Arjovsky et~al.(2017)Arjovsky, Chintala, and
  Bottou]{arjovsky2017wasserstein}
Martin Arjovsky, Soumith Chintala, and L{\'e}on Bottou.
\newblock Wasserstein gan.
\newblock \emph{arXiv preprint arXiv:1701.07875}, 2017.

\bibitem[Bishop(2006)]{bishop2006prml}
Christopher~M. Bishop.
\newblock \emph{Pattern Recognition and Machine Learning (Information Science
  and Statistics)}.
\newblock Springer-Verlag, Berlin, Heidelberg, 2006.
\newblock ISBN 0387310738.

\bibitem[Bregman(1967)]{bregman1967divegence}
Lev~M Bregman.
\newblock The relaxation method of finding the common point of convex sets and
  its application to the solution of problems in convex programming.
\newblock \emph{USSR computational mathematics and mathematical physics},
  7\penalty0 (3):\penalty0 200--217, 1967.

\bibitem[Chen et~al.(2018)Chen, Dai, Pu, Zhou, Li, Su, Chen, and
  Carin]{chen2018symmetric}
Liqun Chen, Shuyang Dai, Yunchen Pu, Erjin Zhou, Chunyuan Li, Qinliang Su,
  Changyou Chen, and Lawrence Carin.
\newblock Symmetric variational autoencoder and connections to adversarial
  learning.
\newblock In \emph{International Conference on Artificial Intelligence and
  Statistics}, pp.\  661--669, 2018.

\bibitem[Dempster et~al.(1977)Dempster, Laird, and Rubin]{dempster1977em}
Arthur~P Dempster, Nan~M Laird, and Donald~B Rubin.
\newblock Maximum likelihood from incomplete data via the em algorithm.
\newblock \emph{Journal of the royal statistical society. Series B
  (methodological)}, pp.\  1--38, 1977.

\bibitem[Goodfellow et~al.(2014)Goodfellow, Pouget-Abadie, Mirza, Xu,
  Warde-Farley, Ozair, Courville, and Bengio]{goodfellow2014gan}
Ian Goodfellow, Jean Pouget-Abadie, Mehdi Mirza, Bing Xu, David Warde-Farley,
  Sherjil Ozair, Aaron Courville, and Yoshua Bengio.
\newblock Generative adversarial nets.
\newblock In \emph{Advances in neural information processing systems}, pp.\
  2672--2680, 2014.

\bibitem[Hiriart-Urruty \& Lemar{\'e}chal(2012)Hiriart-Urruty and
  Lemar{\'e}chal]{hiriart2012conv_ana}
Jean-Baptiste Hiriart-Urruty and Claude Lemar{\'e}chal.
\newblock \emph{Fundamentals of convex analysis}.
\newblock Springer Science \& Business Media, 2012.

\bibitem[Jacobs et~al.(1991)Jacobs, Jordan, Nowlan, and Hinton]{jacobs1991emm}
Robert~A Jacobs, Michael~I Jordan, Steven~J Nowlan, and Geoffrey~E Hinton.
\newblock Adaptive mixtures of local experts.
\newblock \emph{Neural computation}, 3\penalty0 (1):\penalty0 79--87, 1991.

\bibitem[Kakade(2002)]{kakade2002comp_feat}
Sham~M Kakade.
\newblock A natural policy gradient.
\newblock In \emph{Advances in neural information processing systems}, pp.\
  1531--1538, 2002.

\bibitem[Kingma \& Ba(2014)Kingma and Ba]{kingma2014adam}
Diederik Kingma and Jimmy Ba.
\newblock Adam: A method for stochastic optimization.
\newblock \emph{International Conference on Learning Representations}, 12 2014.

\bibitem[Kingma \& Welling(2013)Kingma and Welling]{kingma2013vae}
Diederik~P Kingma and Max Welling.
\newblock Auto-encoding variational bayes.
\newblock \emph{arXiv preprint arXiv:1312.6114}, 2013.

\bibitem[Kullback \& Leibler(1951)Kullback and Leibler]{kullback1951kl}
Solomon Kullback and Richard~A Leibler.
\newblock On information and sufficiency.
\newblock \emph{The annals of mathematical statistics}, 22\penalty0
  (1):\penalty0 79--86, 1951.

\bibitem[Li et~al.(2019)Li, Bai, Li, Wang, Chen, and Carin]{li2019adversarial}
Chunyuan Li, Ke~Bai, Jianqiao Li, Guoyin Wang, Changyou Chen, and Lawrence
  Carin.
\newblock Adversarial learning of a sampler based on an unnormalized
  distribution.
\newblock In \emph{The 22nd International Conference on Artificial Intelligence
  and Statistics}, pp.\  3302--3311, 2019.

\bibitem[Maal{\o}e et~al.(2016)Maal{\o}e, S{\o}nderby, S{\o}nderby, and
  Winther]{maaloe2016auxiliary}
Lars Maal{\o}e, Casper~Kaae S{\o}nderby, S{\o}ren~Kaae S{\o}nderby, and Ole
  Winther.
\newblock Auxiliary deep generative models.
\newblock In \emph{International Conference on Machine Learning}, pp.\
  1445--1453, 2016.

\bibitem[Nguyen et~al.(2010)Nguyen, Wainwright, and
  Jordan]{nguyen2010fganbound}
XuanLong Nguyen, Martin~J Wainwright, and Michael~I Jordan.
\newblock Estimating divergence functionals and the likelihood ratio by convex
  risk minimization.
\newblock \emph{IEEE Transactions on Information Theory}, 56\penalty0
  (11):\penalty0 5847--5861, 2010.

\bibitem[Nowozin et~al.(2016)Nowozin, Cseke, and Tomioka]{nowozin2016fgan}
Sebastian Nowozin, Botond Cseke, and Ryota Tomioka.
\newblock f-gan: Training generative neural samplers using variational
  divergence minimization.
\newblock In \emph{Advances in Neural Information Processing Systems}, pp.\
  271--279, 2016.

\bibitem[Opper \& Saad(2001)Opper and Saad]{opper2001meanField}
Manfred Opper and David Saad.
\newblock \emph{Advanced mean field methods: Theory and practice}.
\newblock MIT press, 2001.

\bibitem[Pentland \& Liu(1999)Pentland and Liu]{Pentland1999}
Alex Pentland and Andrew Liu.
\newblock Modeling and prediction of human behavior.
\newblock \emph{Neural Comput.}, 11\penalty0 (1):\penalty0 229--242, January
  1999.
\newblock ISSN 0899-7667.
\newblock \doi{10.1162/089976699300016890}.
\newblock URL \url{http://dx.doi.org/10.1162/089976699300016890}.

\bibitem[Poole et~al.(2016)Poole, Alemi, Sohl-Dickstein, and
  Angelova]{poole2016iGenObj}
Ben Poole, Alexander~A Alemi, Jascha Sohl-Dickstein, and Anelia Angelova.
\newblock Improved generator objectives for gans.
\newblock \emph{arXiv preprint arXiv:1612.02780}, 2016.

\bibitem[Ranganath et~al.(2016)Ranganath, Tran, and
  Blei]{ranganath2016hierarchical}
Rajesh Ranganath, Dustin Tran, and David Blei.
\newblock Hierarchical variational models.
\newblock In \emph{International Conference on Machine Learning}, pp.\
  324--333, 2016.

\bibitem[Robicquet et~al.(2016)Robicquet, Sadeghian, Alahi, and
  Savarese]{Robicquet2016}
Alexandre Robicquet, Amir Sadeghian, Alexandre Alahi, and Silvio Savarese.
\newblock Learning social etiquette: Human trajectory understanding in crowded
  scenes.
\newblock volume 9912, pp.\  549--565, 10 2016.
\newblock ISBN 978-3-319-46483-1.
\newblock \doi{10.1007/978-3-319-46484-8_33}.

\bibitem[Srivastava et~al.(2014)Srivastava, Hinton, Krizhevsky, Sutskever, and
  Salakhutdinov]{srivastava2014dropout}
Nitish Srivastava, Geoffrey Hinton, Alex Krizhevsky, Ilya Sutskever, and Ruslan
  Salakhutdinov.
\newblock Dropout: a simple way to prevent neural networks from overfitting.
\newblock \emph{The Journal of Machine Learning Research}, 15\penalty0
  (1):\penalty0 1929--1958, 2014.

\bibitem[Sugiyama et~al.(2012)Sugiyama, Suzuki, and
  Kanamori]{sugiyama2012bregmandre}
Masashi Sugiyama, Taiji Suzuki, and Takafumi Kanamori.
\newblock Density-ratio matching under the bregman divergence: a unified
  framework of density-ratio estimation.
\newblock \emph{Annals of the Institute of Statistical Mathematics},
  64\penalty0 (5):\penalty0 1009--1044, 2012.

\bibitem[Uehara et~al.(2016)Uehara, Sato, Suzuki, Nakayama, and
  Matsuo]{uehara2016bgan}
Masatoshi Uehara, Issei Sato, Masahiro Suzuki, Kotaro Nakayama, and Yutaka
  Matsuo.
\newblock Generative adversarial nets from a density ratio estimation
  perspective.
\newblock \emph{arXiv preprint arXiv:1610.02920}, 2016.

\end{thebibliography}
\bibliographystyle{iclr2020_conference}

\newpage
\appendix
\section{Pseudo Code \label{ap:code}}
\begin{algorithm}
    \SetKwInOut{Input}{Input}
    \underline{EIM-for-GMMs}$(\lbrace \cvec{x}_\textrm{p}^{(j)} \rbrace_{j = 1 \cdots N}, q(\cvec{x}))$\;
    \Input{Data $\lbrace \cvec{x}_\textrm{p}^{(j)} \rbrace_{j = 1 \cdots N}$, Initial Model $q(\cvec{x}) = \sum_{i=1}^d q(\cvec{x}|z_i) q(z_i) = \sum_{i=1}^d \pi_i \mathcal{N}(\cvec{x} | \cvec{\mu}_i, \cmat{\Sigma}_i)$}
    \For{$i$ in number of iterations}
	{\textbf{E-Step:} \\
     $\old{q}(z) = q(z)$, $ \old{q}(\cvec{x} | z_i) = q(\cvec{x}|z_i)$ for all components $i$ \\
     \textbf{Update Density Ratio Estimator:}\\
     sample data from model $\lbrace \cvec{x}_\textrm{q}^{(j)} \rbrace_{j = 1 \cdots N} \sim \old{q}(\cvec{x})$ \\
	 retrain density ratio estimator $\phi(\cvec{x})$	on $\lbrace \cvec{x}_\textrm{p}^{(j)} \rbrace_{j = 1 \cdots N}$ and $\lbrace \cvec{x}_\textrm{q}^{(j)} \rbrace_{j = 1 \cdots N}$ \\
	 \textbf{M-Step Coefficients:} \\
	 \For {$i$ in number of components}{
	 	compute loss $l_i = \dfrac{1}{N} \sum_{j=1}^N \phi \left(\cvec{x}_\textrm{q}^{(j)} \right)$ with samples  $\lbrace \cvec{x}_\textrm{q}^{(j)} \rbrace_{j = 1 \cdots N} \sim \old{q}(\cvec{x}|z_i)$
	 	}
	 update $q(z)$ using losses $l_i$ and MORE equations \\
	 \textbf{M-Step Components:} \\
	 \For {$i$ in number of components}{
	 	fit $\hat{\phi}(\cvec{x})$ surrogate to pairs $\left(\cvec{x}_\textrm{q}^{(j)}, \phi\left(\cvec{x}_\textrm{q}^{(j)}\right)\right)$ with samples $\lbrace \cvec{x}_\textrm{q}^{(j)} \rbrace_{j = 1 \cdots N} \sim \old{q}(\cvec{x}|z_i)$ \\
	 	update $q(\cvec{x}|z_i)$ using surrogate $\hat{\phi}(\cvec{x})$ and MORE equations
	 }	 
    }
    \caption{Expected Information Maximization for Gaussian Mixture Models.}
    \label{algo:eim_gmm}
\end{algorithm}

Pseudo-code for EIM for GMMs can be found in \autoref{algo:eim_gmm}

\section{Derivations \label{ap:bound}}
Derivations of the upper bound stated in \autoref{eq:upperbound}.
We assume latent variable models $q(x) = \int q(x|z)q(z)dz$ and use the identities $q(x|z)q(z) = q(z|x)q(x)$ and $\log q(x) = \log q(x|z)q(z) - \log q(z|x)$. 
\begin{align*}
 & \KL{q(x)}{p(x)} = \int q(x) \log \dfrac{q(x)}{p(x)} dx = \iint q(x|z)q(z) \log \dfrac{q(x)}{p(x)} dzdx \\
=& \iint q(x|z)q(z) \left( \log \dfrac{q(x|z)q(z)}{p(x)} - \log q(z|x) \right) dzdx \\
=& \iint q(x|z)q(z) \left( \log \dfrac{q(x|z)q(z)}{p(x)} - \log q(z|x) + \log \tilde{q}(z|x) - \log \tilde{q}(z|x) \right) dzdx \\
=& \iint q(x|z)q(z) \left( \log \dfrac{q(x|z)q(z)}{p(x)} - \log \tilde{q}(z|x) \right) dz dx - \iint q(x|z)q(z) \left( \log q(z|x) - \log \tilde{q}(z|x) \right) dz dx \\
=& \iint q(x|z)q(z) \left( \log \dfrac{q(x|z)q(z)}{p(x)} - \log \tilde{q}(z|x) \right) dz dx - \int q(x) \int q(z|x) \log \dfrac{q(z|x)}{\tilde{q}(z|x)} dz dx \\
=& U(q, \tilde{q}, p) - \mathbb{E}_{q(x)} \left[ \KL{q(z|x)}{\tilde{q}(z|x)} \right]. 
\end{align*}
After plugging the E-Step, i.e., $\tilde{q}(z|x) = \old{q}(x|z) \old{q}(z) / \old{q}(x)$, into the objective it simplifies to 
\begin{align*}
&U(q, \tilde{q}, p) \\
=& \iint q(x|z)q(z) \left( \log \dfrac{q(x|z)q(z)}{p(x)} - \log \dfrac{\old{q}(x|z) \old{q}(z)}{\old{q}(x)} \right) dz dx \\
=& \iint q(x|z)q(z) \left( \log q(x|z) + \log q(z) - \log p(x) - \log \old{q}(x|z) - \log \old{q}(z) + \log \old{q}(x) \right) dz dx \\
=& \iint q(x|z)q(z) \left( \log \dfrac{\old{q}(x)}{p(x)} + \log \dfrac{q(x|z)}{\old{q}(x|z)} + \log \dfrac{q(z)}{\old{q}(z)} \right) dz dx \\
=& \int q(z) \left( \int q(x|z) \left( \log \dfrac{\old{q}(x)}{p(x)} + \log \dfrac{q(x|z)}{\old{q}(x|z)} \right) dx + \log \dfrac{q(z)}{\old{q}(z)} \right) dz \\
=& \int q(z) \int q(x|z) \log \dfrac{\old{q}(x)}{p(x)} dxdz + \int q(z) \int q(x|z) \log \dfrac{q(x|z)}{\old{q}(x|z)} dxdz + \int q(z) \log \dfrac{q(z)}{\old{q}(z)} dz \\
=& \iint q(x|z)q(z) \log \dfrac{\old{q}(x)}{p(x)} dzdx + \mathbb{E}_{q(z)} \left[\KL{q(x|z)}{\old{q}(x|z)} \right] + \KL{q(z)}{\old{q}(z)},
\end{align*}
which concludes the derivation of upper bound of latent variable models.
\subsection{Derivations Conditional Upper Bound}
\label{a:sec:cond_upper_bound}
By introducing an auxiliary distribution $\tilde{q}(z|x,y)$ the upper bound to the expected KL for conditional latent variable models $q(x|y) = \int q(x|z,y) q(z|y) dz$ can be derived by
\begin{align*}
 & \mathbb{E}_{p(y)}\left[\KL{q(x|y)}{p(x|y)}\right] = \iint p(y) q(x|y) \log \dfrac{q(x|y)}{p(x|y)} dx dy \\
=& \int p(y) \iint q(x|z,y)q(z|y) \left( \log \dfrac{q(x|z,y)q(z|y)}{p(x|y)} - \log q(z|x, y) \right) dzdxdy \\
=& \int p(y) \iint q(x|z,y)q(z|y) \\ 
 &\cdot \left( \log \dfrac{q(x|z,y)q(z|y)}{p(x|y)} - \log q(z|x, y) + \log \tilde{q}(z|x, y) - \log \tilde{q}(z|x, y) \right) dzdxdy \\
=& \int p(y) \iint q(x|z,y)q(z|y) \left( \log \dfrac{q(x|z,y)q(z|y)}{p(x|y)} - \log \tilde{q}(z|x, y) \right) dzdxdy  \\ 
 &- \int p(y) \iint q(x|z, y)q(z, y) \left( \log q(z|x, y) - \log \tilde{q}(z|x, y) \right) dzdxdy \\
=& \int p(y) \iint q(x|z,y)q(z|y) \left( \log \dfrac{q(x|z,y)q(z|y)}{p(x|y)} - \log \tilde{q}(z|x, y) \right) dzdxdy  \\ 
 &- \iint p(y) q(x|y) \int q(z|x, y) \log \dfrac{q(z|x, y)}{\tilde{q}(z|x,y)} dz dx dy  \\
=& U(q, \tilde{q}, p) - \mathbb{E}_{p(y), q(x|y)} \left[ \KL{q(z|x, y)}{\tilde{q}(z|x,y)} \right].  
\end{align*}
During the E-step the bound is tightened by setting  $\tilde{q}(z|x,y) = \old{q}(x|z,y)\old{q}(z|y)/ \old{q}(x|y)$.
\begin{align*}
&U(q, \tilde{q}, p) \\
=& \int p(y) \iint q(x|z,y)q(z|y) \left( \log \dfrac{q(x|z,y)q(z|y)}{p(x|y)} - \log \dfrac{\old{q}(x|z, y) \old{q}(z|y)}{\old{q}(x|y)} \right) dz dx dy \\
=& \int p(y) \iint q(x|z,y)q(z|y) \\
 &\cdot \left( \log q(x|z,y) + \log q(z|y) - \log p(x|y) - \log \old{q}(x|z,y) - \log \old{q}(z|y) + \log \old{q}(x|y) \right) dz dx dy \\
=& \int p(y) \iint q(x|z,y)q(z|y) \left( \log \dfrac{\old{q}(x|y)}{p(x|y)} + \log \dfrac{q(x|z,y)}{\old{q}(x|z,y)} + \log \dfrac{q(z|y)}{\old{q}(z|y)} \right) dz dx dy\\
=& \int p(y) \int q(z|y) \left( \int q(x|z,y) \left( \log \dfrac{\old{q}(x|y)}{p(x|y)} + \log \dfrac{q(x|z,y)}{\old{q}(x|z,y)} \right)dx + \log \dfrac{q(z|y)}{\old{q}(z|y)} \right) dz  dy\\
=& \int p(y) \int q(z|y) \int q(x|z,y) \log \dfrac{\old{q}(x|y)}{p(x|y)} dxdzdy \\
 &+ \int p(y) \int q(z|y) \int q(x|z,y) \log \dfrac{q(x|z,y)}{\old{q}(x|z,y)} dxdzdy + \int p(y) \int q(z|y) \log \dfrac{q(z|y)}{\old{q}(z|y)} dzdy \\
=& \iiint p(y)q(z|y)q(x|z,y) \log \dfrac{\old{q}(x|y)}{p(x|y)} dxdzdy \\
 &+ \mathbb{E}_{p(y), q(z|y)} \left[ \KL{q(x|z,y)}{\old{q}(x|z,y)} \right] + \mathbb{E}_{p(y)} \left[ \KL{q(z|y)}{\old{q}(z|y)} \right],
\end{align*}
which concludes the derivation of the upper bound for conditional latent variable models.

\subsection{Using MORE for closed form updates for GMMs \label{ap:VIPS}}
The MORE algorithm, as introduced by \cite{abdolmaleki2015more}, can be used to solve optimization problems of the following form
\begin{align*}
&\textrm{argmax}_{q(x)} \mathbb{E}_{q(x)}[f(x)] \quad \textrm{s.t.} \quad \KL{q(x)}{q_\textrm{old}(x)} \leq \epsilon
\end{align*}
for an exponential family distribution $q(x)$, some function $f(x)$, and an upper bound on the allowed change, $\epsilon$.
\cite{abdolmaleki2015more} show that the optimal solution is given by 
\begin{align*}
q(x) \propto q_\textrm{old}(x) \exp \left(\dfrac{f(x)}{\eta} \right) = \exp\left(\dfrac{\eta \log q_\textrm{old}(x) + f(x)}{\eta} \right),
\end{align*}
where $\eta$ denotes the Lagrangian multiplier corresponding to the KL constraint.
In order to obtain this Lagrangian multiplier, the following, convex, dual function has to be minimized
\begin{align}
g(\eta) = \eta \epsilon + \eta \log \int \exp \left( \dfrac{\eta \log q_\textrm{old}(x) + f(x)}{\eta } \right) dx.\label{eq:more_dual}
\end{align}  

For discrete distributions, such as the categorical distribution used to represent the coefficients of a GMM, we can directly work with those equations. 
For continuous distributions, \cite{abdolmaleki2015more} propose approximating $f(x)$ with a local surrogate.
The features to fit this surrogate are chosen such that they are compatible \citep{kakade2002comp_feat}, i.e., of the same form as the distributions sufficient statistics.
For multivariate Gaussians, the sufficient statistics are squared features and thus the surrogate compatible to such a Gaussian distribution is given by 
$$\hat{f}(x) = -\dfrac{1}{2} \cvec{x}^T \cvec{F} \cvec{x} + \hat{\cvec{f}}^T \cvec{x} + f_0.$$
The parameters of this surrogate can now be used to update the natural parameters of the Gaussian, i.e, the precision matrix $\cmat{Q} = \cmat{\Sigma}^{-1}$ and $\cvec{q} = \cmat{\Sigma}^{-1}\cvec{\mu}$ by
\begin{align*}
\cmat{Q} =\old{\cmat{Q}} +   \dfrac{1}{\eta} \hat{\cmat{F}} \quad \text{and} \quad \cvec{q} =  \old{\cvec{q}} +  \dfrac{1}{\eta } \hat{\cvec{f}}.
\end{align*}

In order to apply the MORE algorithm to solve the optimization problems stated in \autoref{eq:EIM_weight} and \autoref{eq:EIM_component} we make two trivial modifications. First, we invert the signs in \autoref{eq:EIM_weight} and \autoref{eq:EIM_component}, as we are now maximizing. 
Second, to account for the additional KL term in our objectives, we add $1$ to $\eta$, everywhere except the first term of the sum in \autoref{eq:more_dual}.

\section{Elaboration on Related Work}
\subsection{Relation between EIM and EM \label{ap:EM}}
Recall that the Expectation-Maximization (EM) algorithm \citep{dempster1977em} maximizes the log-likelihood of the data by iteratively maximizing and tightening the following lower bound 
\begin{align*}
  &\mathbb{E}_{p(x)} \left[ \log q(x)  \right] = \mathbb{E}_{p(x)} \left[ \int \tilde{q}(z|x) \log \dfrac{q(x , z)}{\tilde{q}(z|x)} dz\right] +  \mathbb{E}_{p(x)} \left[ \int \tilde{q}(z|x) \log \dfrac{\tilde{q}(z|x)}{q(z | x)} dz \right] \\
=& \underbrace{\mathcal{L}(q, \tilde{q})}_\textrm{lower bound} +  \underbrace{\mathbb{E}_{p(x)} \left[ \KL{\tilde{q}(z|x)}{q(z | x)}\right]}_{\geq 0}.
\end{align*}

It is instructive to compare our upper bound (\autoref{eq:upperbound}) to this lower bound.
As mentioned, maximizing the likelihood is equivalent to minimizing the M-projection, i.e., $\textrm{argmin}_{q(x)}\KL{p(x)}{q(x)}$, where, in relation to our objective, the model and true distribution have switched places in the non-symmetric KL objective.
Like our approach, EM introduces an auxiliary distribution $\tilde{q}(z|x)$ and bounds the objective from below by subtracting the KL between auxiliary distribution and model, i.e., $\KL{\tilde{q}(z|x)}{q(z|x)}$. 
In contrast, we obtain our upper bound by adding $\KL{q(z|x)}{\tilde{q}(z|x)}$ to the objective. 
Again, the distributions have exchanged places within the KL.

\subsection{Equality of $f$ and $b$-GAN \label{ap:GAN}}
As pointed out in \autoref{sec:GAN} both the $f$-GAN \citep{nowozin2016fgan} and the $b$-GAN \citep{uehara2016bgan} yield the same objective for the I-projection. 

We start with the $f$-GAN. \cite{nowozin2016fgan} propose the following adversarial objective, based on a variatonal bound for $f$-divergences \citep{nguyen2010fganbound}
$$\textrm{argmin}_{q(x)} \textrm{argmax}_{V(x)} F(q(x),V(x)) = \mathbb{E}_{p(x)}\left[g(V(x)) \right] - \mathbb{E}_{q(x)} \left[f^*(g(V(x))) \right].$$
Here $V(x)$ denotes a neural network with linear output, $g(v)$ the output activation and $f^*(t)$ the Fenchel conjugate \citep{hiriart2012conv_ana} of $f(u)$, i.e., the generator function of the $f$-divergence. 
For the I-projection $f(u) = - \log u$ and $f^*(t) = - 1 - \log(t)$. 
In theory, the only restriction posed on the choice of $g(v)$ is that it outputs only values within the domain of $f^*(t)$ \cite{nowozin2016fgan} suggest $g$'s for various $f-$ divergences and chose exclusively monotony increasing functions which output large values for samples that are believed to be from the data distribution. 
For the I-projection they suggest $g(v)=-exp(-v)$. Thus the $f$-GAN objective for the I-projection is given by
$$\textrm{argmin}_{q(x)} \textrm{argmax}_{V(x)} F(q(x),V(x)) = -\mathbb{E}_{p(x)}\left[\exp(-V(x) \right] + \mathbb{E}_{q(x)} \left[1 -  V(x) \right].$$

The $b$-GAN objective follows from the density ratio estimation framework given by \cite{sugiyama2012bregmandre} and is given by 
$$\textrm{argmin}_{q(x)} \textrm{argmax}_{r(x)} \mathbb{E}_{p(x)} \left[ f'(r(x)) \right] - \mathbb{E}_{q(x)}  \left[f'(r(x))r(x) - f(r(x)) \right]$$
Here $f'(u)$ denotes the derivative of $f(u)$ and $r(x)$ denotes an density ratio estimator. We need to enforce $r(x) > 0$ for all $x$ to obtain a valid density ratio estimate. In practice this is usually done by learning $r_l(x) = \log r(x)$ instead. Plugging $r_l(x)$, $f(u)$ and $f'(u) = 1 / u$ into the general $b$-GAN objective yields
$$\textrm{argmin}_{q(x)} \textrm{argmax}_{r_l(x)} F(q(x),r_l(x)) = -\mathbb{E}_{p(x)}\left[\exp(-r_l(x)) \right] + \mathbb{E}_{q(x)} \left[1 -  r_l(x) \right].$$
Which is the same objective as the $f$-GAN uses. Yet, $f$-GAN and $b$-GAN objectives are not identical for arbitrary $f$-divergences.

\section{Visualization of Samples}
\subsection{Pedestrian and Traffic Prediction}
\begin{figure}
    \centering
    \begin{minipage}{0.25\textwidth}
    \includegraphics[width=\textwidth]{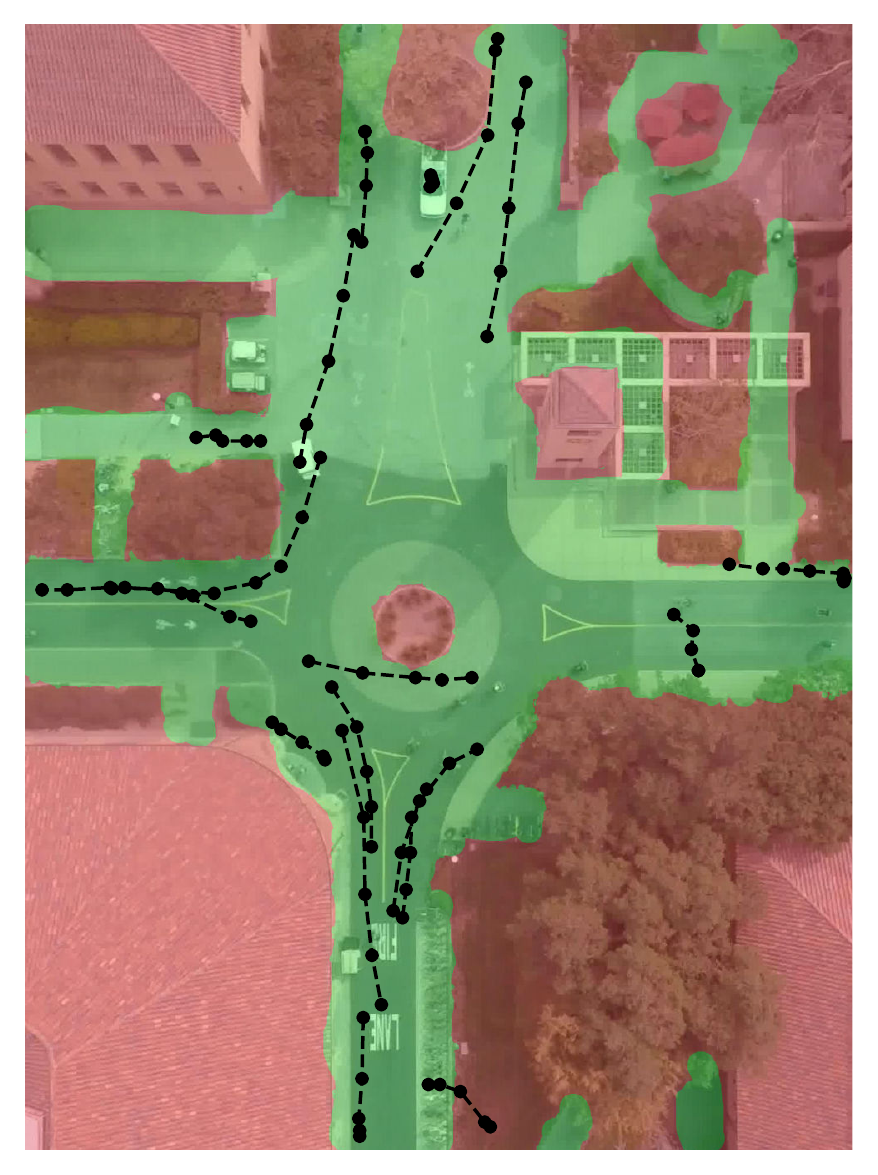}
    \subcaption{Data}
    \end{minipage}%
    \begin{minipage}{0.25\textwidth}
    \includegraphics[width=\textwidth]{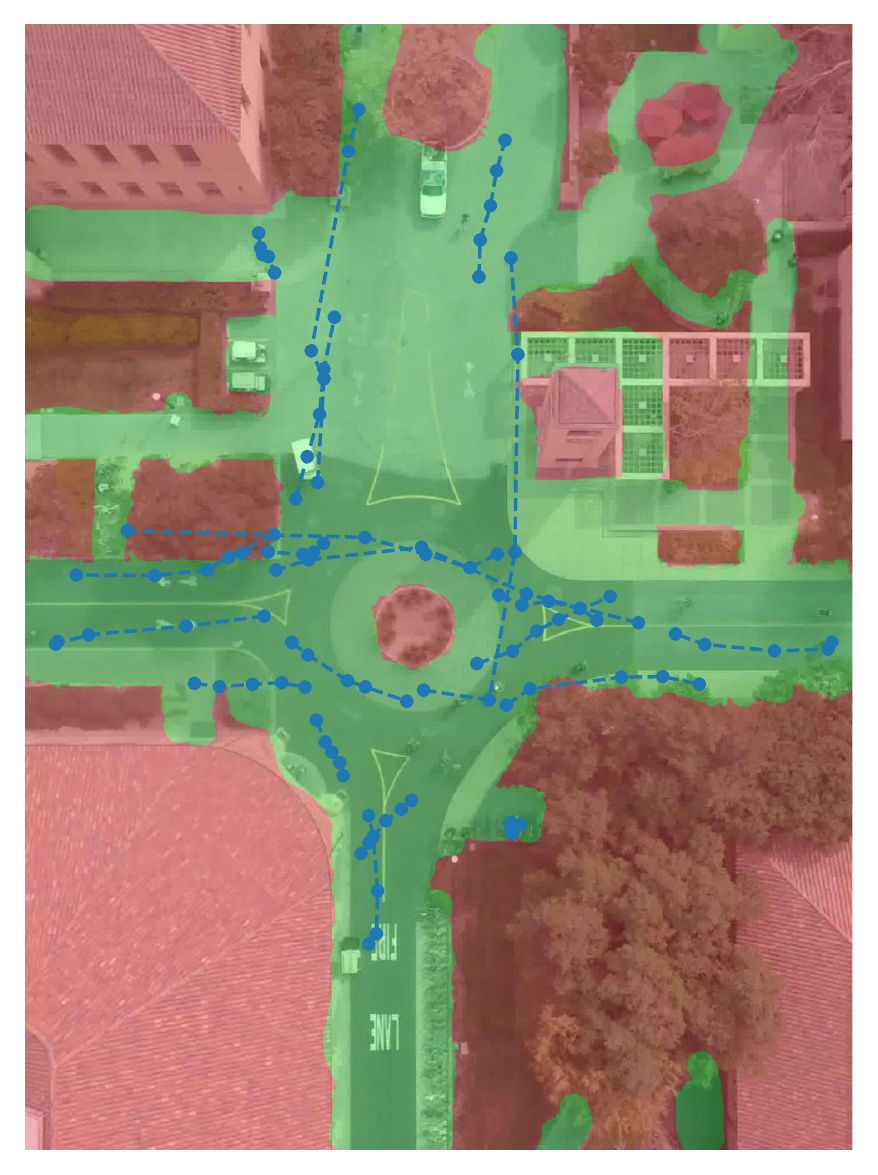}
    \subcaption{EIM}
    \end{minipage}%
    \begin{minipage}{0.25\textwidth}
    \includegraphics[width=\textwidth]{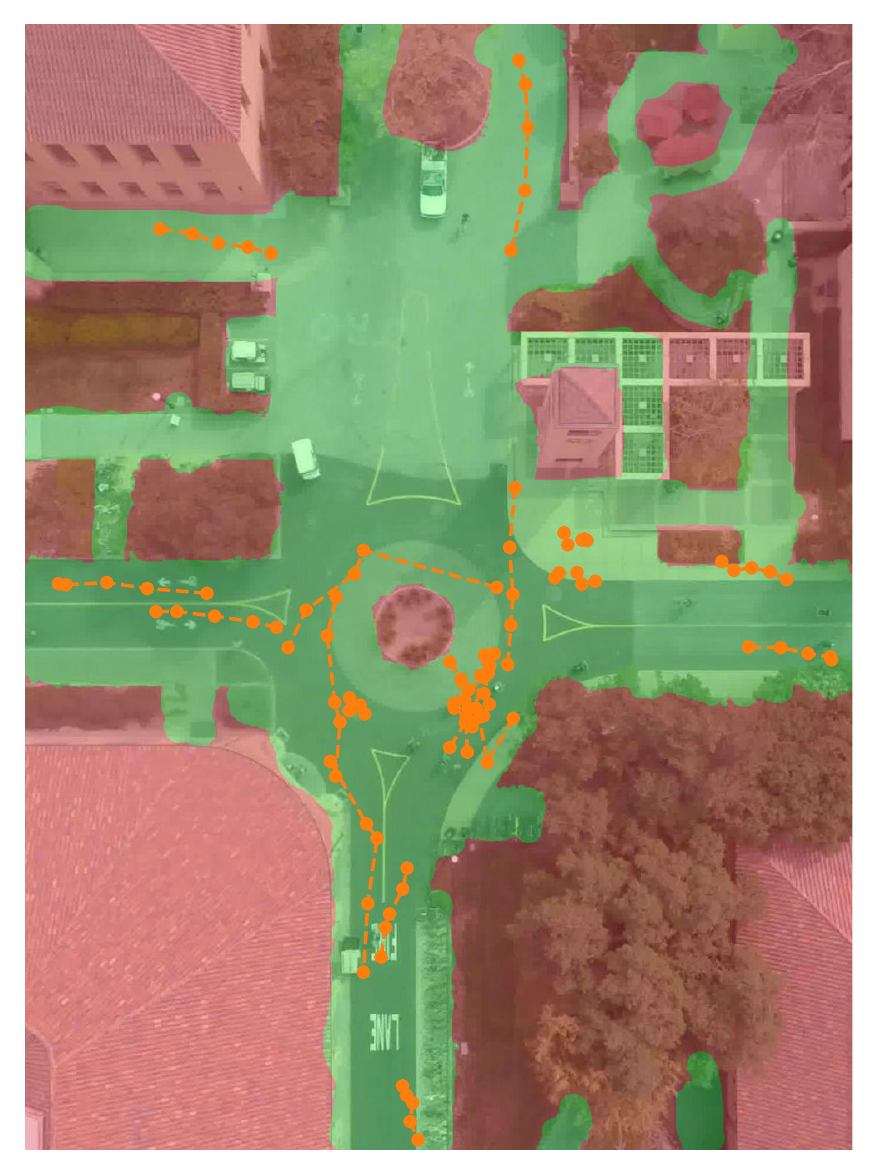}
    \subcaption{EIM + Features}
    \end{minipage}%
    \begin{minipage}{0.25\textwidth}
    \includegraphics[width=\textwidth]{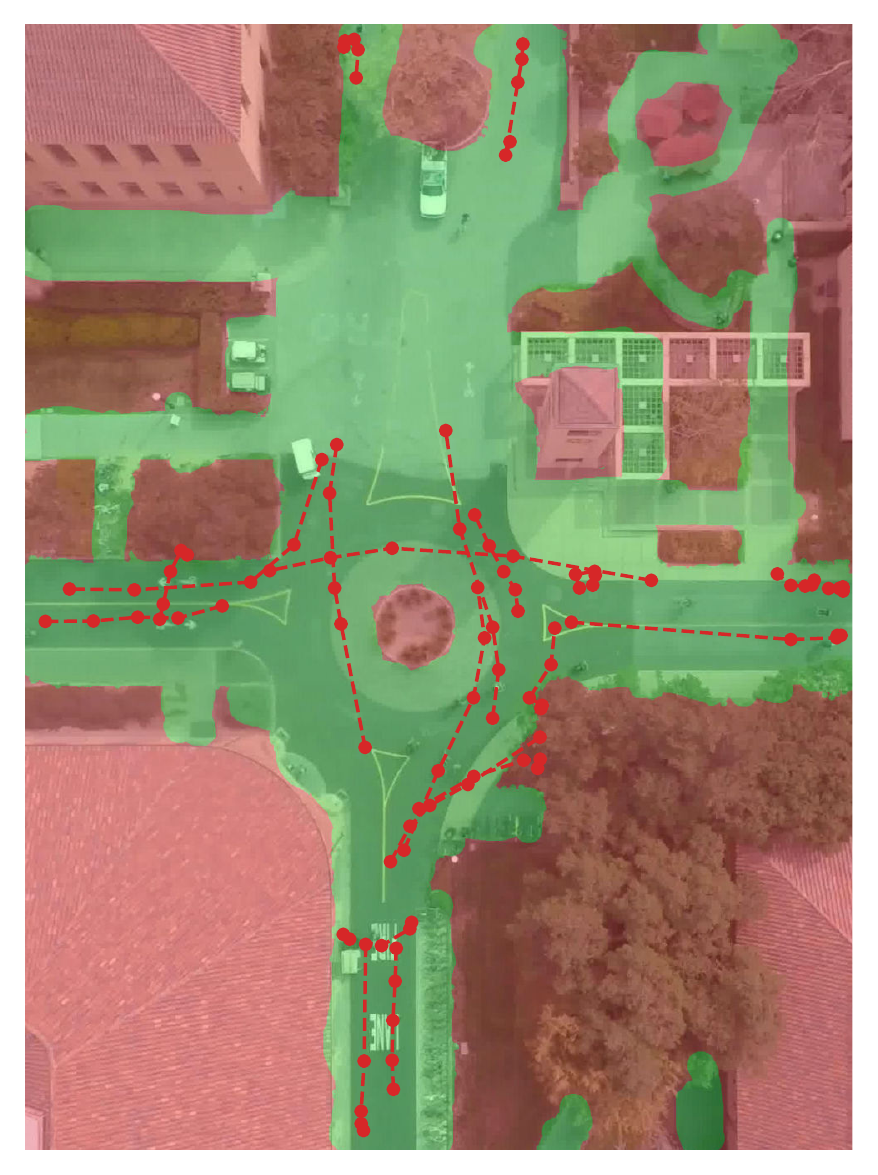}
    \subcaption{EM}
    \end{minipage}
    \caption{Samples from the Dataset, EIM, EIM with features and EM, plotted over the reference image from the Stanford Drone Dataset and the generated mask. In the mask green corresponds to valid regions and red to invalid regions. EIM with the additional feature input generates samples that stay within the 'road mask' and are therefore considered to be more realistic.}
    \label{fig:stanford_samples}
\end{figure}
Samples from the Stanford Drone Dataset can be found in \autoref{fig:stanford_samples}

\section{Hyperparameters \label{ap:experiments}}
In all experiments, we realize the density ratio estimator as fully connected neural networks which we train using Adam \citep{kingma2014adam} and early stopping using a validation set.

\textbf{Comparison to Generative Adversarial Approaches and Ablation Study}
\begin{itemize}
    \item Data: $10,000$ Train Samples, $5,000$ Test Samples, $5,000$ Validation samples (for early stopping the density ratio estimator)
    \item Density Ratio Estimator (EIM) / Variational function $V(x)$ ($f$-GAN): 3 fully connected layers, 50 neurons each, trained with L2 regularization with factor $0.001$, early stopping and batch size $1,000$
    \item Updates EIM: MORE-like updates with $\epsilon = 0.05$ for components and coefficients, $1,000$ samples per component and update
    \item Updates FGAN: Iterate single update steps for generator and discriminator using learning rates of $1e-3$ and batch size of $1,000$.
\end{itemize}
\textbf{Line Reaching with Planar Robot}
\begin{itemize}
    \item Data: $10,000$ train samples, $5,000$ test samples, $5,000$ validation samples (for early stopping the density ratio estimator)
    \item Density Ratio Estimation: 2 fully connected layers of width $100$, early stopping and batch size $1,000$
    \item Updates: MORE-like updates with $\epsilon = 0.005$ for components and coefficients, $1,000$ samples per component and update
\end{itemize}
\textbf{Pedestrian and Traffic Prediction}
\begin{itemize}
    \item Data SDD: $7,500$ train samples, $3,801$ test samples, $3,801$ validation samples (for early stopping the density ratio estimator)
    \item Data NGS: $10,000$ train samples, $5,000$ test samples, $5,000$ validation samples (for early stopping the density ratio estimator)
    \item Density Ratio Estimation: 3 fully connected layers of width $256$, trained with L2     regularization with factor $0.0005$ early stopping and batch size $1,000$.
    \item Updates: MORE-like updates with $\epsilon = 0.01$ for components and coefficients, $1,000$ samples per component and update
\end{itemize}
\textbf{Obstacle Avoidance}
\begin{itemize}
    \item Data: $1,000$ train contexts with $10$ samples each, $500$ test contexts with $10$ samples each, $500$ validation contexts with $10$ samples each (for early stopping the density ratio estimator).
    \item Density Ratio Estimation: 3 fully connected layers of width $256$, trained with L2 regularization with factor $0.0005$ early stopping and batch size $1,000$.
    \item Component and Gating networks: 2 fully connected layer of width $64$ for each component and the gating. Trained with Adam ($\alpha=1e-3, \beta_0 = 0.5$) for $10$ epochs in each iteration.
\end{itemize}

\end{document}